\pdfoutput=1

\documentclass[11pt]{article}

\usepackage[]{acl}

\usepackage{times}
\usepackage{latexsym}

\usepackage[T1]{fontenc}

\usepackage[utf8]{inputenc}

\usepackage{microtype}

\usepackage{amsmath}
\usepackage{amssymb}
\usepackage{framed} 
\usepackage{subcaption}  
\usepackage{tabularx}  
\usepackage{booktabs}  
\usepackage{multirow}  

\makeatletter
\def\new@fontshape{}
\makeatother
\usepackage{siunitx} 

\usepackage{xspace}  

\usepackage{tikz}  
\usetikzlibrary{shapes.geometric, arrows}
\usetikzlibrary{calc, positioning}  
\usepackage{forest}  
\useforestlibrary{linguistics}
\forestapplylibrarydefaults{linguistics}

\usepackage{stmaryrd} 

\usepackage[compress,sort,capitalise]{cleveref}  
\usepackage{hyperref}

\usepackage[inline]{enumitem}  

\usepackage{gb4e}  
\noautomath

\newcommand{\todo}[1]{\textcolor{red}{(\textbf{#1})}}

\newcommand{\sortof}[1]{`#1'}
\newcommand{\lform}[1]{\texttt{#1}}
\newcommand{\graphedge}[1]{`#1'}
\newcommand{\graphnode}[1]{`#1'}
\newcommand{\word}[1]{``#1''}

\newcommand{\sentence}[1]{``#1''}

\crefname{xnumi}{}{}
\creflabelformat{xnumi}{(#2#1#3)}
\crefrangeformat{xnumi}{(#3#1#4)--(#5#2#6)}

\tikzstyle{nnode} = [ellipse, text centered, draw=black, inner sep=2pt] 
\tikzstyle{dnode} = [rectangle, rounded corners, text centered, draw=black] 
\tikzstyle{arrow} = [thick,->,>=stealth]

\definecolor{colorwant}{HTML}{ABBEF1} 
\definecolor{colorgo}{HTML}{DE8BB3} 
\definecolor{colorthe}{HTML}{F1AD83} 
\definecolor{colorboy}{HTML}{FFE19D} 

\newcommand{\app}[1]{\text{\textsc{App}}\ensuremath{_{#1}}\xspace}
\newcommand{\modify}[1]{\text{\textsc{mod}}\ensuremath{_{#1}}\xspace}

\newcommand{\src}[1]{\text{\textsc{#1}}\xspace} 
\newcommand{\srcr}[1]{\textcolor{red}{\src{#1}}} 

\pgfdeclarelayer{background}
\pgfdeclarelayer{main}
\pgfsetlayers{background,main}
\newcommand*{\minWidth}{25}  
\newcommand*{\maxValue}{100}
\newcommand{\makeBar}[3]{%
  \tikz[baseline]{
    \node[anchor=base,text width=\minWidth,align=#2,inner sep=0pt,inner xsep=4pt,outer sep=0pt] (n) {\strut{\hfill#1}};  
    \begin{pgfonlayer}{background}
        {
       \edef\color{#3}
       \pgfmathparse{abs(#1/\maxValue)}
       \edef\contents{{\pgfmathresult}}
       \fill[font=\boldmath,color=\color] (n.north west) rectangle ($(n.south west)!{{\contents}}!(n.south east)$);  
       \fill[font=\boldmath,color=cyan!10] ($(n.north west)!{{\contents}}!(n.north east)$) rectangle (n.south east);  
       }
    \end{pgfonlayer}
  }
}
\newcommand{\asbar}[1]{\makeBar{#1}{center}{cyan!25}}
\newcommand{\stderr}[1]{\tiny$\pm$#1}

\newcommand{\genclass}[1]{\textsc{#1}\xspace} %
\newcommand{\lexg}{\genclass{Lex}}
\newcommand{\propg}{\genclass{Prop}}
\newcommand{\structg}{\genclass{Struct}}

\newcommand{\amBert}{AM+B} 
\newcommand{\amToken}{AM} 
\newcommand{\dist}{dist\xspace} 

\newcommand{\oldNumber}[1]{\textcolor{black!40}{#1}}

\usepackage{color}
\usepackage{bm}
\definecolor{orange}{rgb}{1,0.5,0}
\definecolor{mdgreen}{rgb}{0.05,0.6,0.05}
\definecolor{mdblue}{rgb}{0,0,0.7}
\definecolor{dkblue}{rgb}{0,0,0.5}
\definecolor{dkgray}{rgb}{0.3,0.3,0.3}
\definecolor{slate}{rgb}{0.25,0.25,0.4}
\definecolor{gray}{rgb}{0.5,0.5,0.5}
\definecolor{ltgray}{rgb}{0.7,0.7,0.7}
\definecolor{purple}{rgb}{0.7,0,1.0}
\definecolor{lavender}{rgb}{0.65,0.55,1.0}
\definecolor{brown}{rgb}{0.6,0.2,0.2}

\newcommand{\code}[1]{}

\title{Compositional Generalization Requires Compositional Parsers}

\author{Pia Wei{\ss}enhorn\thanks{\ \ denotes equal contribution} \and Yuekun Yao\footnotemark[1] \and Lucia Donatelli \and Alexander Koller\\
  Department of Language Science and Technology\\
  Saarland University, Germany \\
  \texttt{\{piaw, ykyao, donatelli, koller\}@coli.uni-saarland.de}}

\begin{document}
\maketitle
\begin{abstract}
  A rapidly growing body of research on
  \textit{compositional generalization} investigates the ability of a semantic
  parser to dynamically recombine linguistic elements seen in
  training into unseen sequences. We present a systematic comparison of sequence-to-sequence
  models and models guided by compositional principles on the
  recent COGS corpus \citep{kim-linzen-2020-cogs}. Though seq2seq models can perform well on lexical tasks, they perform with near-zero accuracy on \textit{structural generalization}
  tasks that require novel syntactic structures; this holds true even when they are trained to predict syntax instead of semantics. In contrast, compositional models achieve near-perfect accuracy on structural generalization; we present new results confirming this from the AM parser \cite{groschwitz-etal-2021-learning}. Our findings show structural generalization is a key measure of compositional generalization and requires models that are aware of complex structure. 
  
  
\end{abstract}


\section{Introduction}\label{sec:introduction}

Compositionality is a fundamental principle of natural language
semantics: ``The meaning of a whole [expression] is a function of the
meanings of the parts and of the way they are syntactically combined''
\citep{partee-1984-compositionality}. 
 
A growing body of research focuses on
\emph{compositional generalization}, the ability of a
semantic parser to
combine known linguistic elements in novel structures in ways akin to humans. For example,
observing the meanings of \textit{``The hedgehog ate a cake''} and \textit{``A baby
liked the penguin,''} can a model predict the meaning of \textit{``A baby liked
the hedgehog''}?
Dynamic, compositional recombination helps explain efficient human language learning and usage, and investigating whether NLP models make use of the same property offers important insight into their behavior. 

Current research on compositional generalization shows
the task to be challenging and complex. Such research centers around a
number of corpora designed specifically for the task, including SCAN
\citep{lake-baroni-2018-generalization} and CFQ
\citep{keysers-etal-2020-measuring}. We focus on COGS \cite{kim-linzen-2020-cogs}, a synthetic
semantic parsing corpus of English whose test set consists of 21
\emph{generalization types} such as the example above
(\cref{sec:background}). Kim and Linzen report that simple
sequence-to-sequence (seq2seq) models such as LSTMs and Transformers
struggle with many of their generalization types, achieving an overall
highest accuracy on the generalization set of 35\%. 
Subsequent work
has improved accuracy on the COGS generalization set considerably
\citep{tay-etal-2021-pretrained,akyurek-andreas-2021-lexicon,conklin-etal-2021-meta,csordas-etal-2021-devil,orhan-2021-compgen,zheng-lapata-2021-disentangled},
but the accuracy of even the best seq2seq models remains below 88\%.
By contrast, \citet{liu-etal-2021-learning} report an accuracy of
98\%, using an algebraic model that implements compositionality
(\cref{sec:related_work}).

Here, we investigate whether this difference in compositional
generalization accuracy is incidental, or whether there is a
systematic difference between seq2seq models and models that are
guided by compositional principles and aware of complex
structure. Comparisons between entire classes of models must be made
with care. Thus in order to make claims about the class of
compositional models, we first work out a second compositional model
for COGS (in addition to Liu et al.'s). We apply the AM
parser \cite{groschwitz-etal-2021-learning}, a compositional semantic
parser which can parse a variety of graphbanks fast and accurately
\cite{lindemann-etal-2020-fast}, to COGS after minimal adaptations
(\cref{sec:amparsing4cogs}). The AM parser achieves a generalization
accuracy above 98\%, making it the first semantic parser shown to
perform accurately on both COGS
and broad-coverage semantic parsing.

We then
compare these two
compositional models to all published seq2seq models for COGS. We find
that the difference in generalization accuracy can be attributed
specifically to \emph{structural} types of compositional
generalization, which require the parser to generalize to novel
syntactic structures that were not observed in training. While the
compositional parsers achieve excellent accuracy on these
generalization types, all known seq2seq models perform very poorly,
with accuracies close to zero. This is even true for BART
\cite{lewis-etal-2020-bart-acl}, which we apply to COGS for the first
time; this is surprising because BART achieves very high accuracy on
broad-coverage semantic parsing tasks \cite{bevilacqua-etal-2021-one}. We conclude that seq2seq
models, as a class, seem to have a weakness with regard to structural
generalization that compositional models overcome
(Section~\ref{subsec:first:experiments}).

Finally, we investigate the role of syntax in compositional
generalization (\cref{sec:analysis-extensions}). We show that parsers
which explicitly model syntactic tree structures can easily learn
structural generalization when trained to predict syntax trees on
COGS, whereas BART again performs poorly. 
BART does
not learn structural generalization even if we enrich its input with
syntactic information. Thus, the poor performance of seq2seq models on
structural generalization is not specifically due to representational
choices in COGS, or even to the specific compositional demands of
semantic parsing; structural generalization requires
structure-aware models.

We discuss implications for future work on compositional
generalization in \cref{sec:discussion}. All code will be made
publicly available upon acceptance.

\newcommand{\ignore}[1]{}
\ignore{

Our contributions: \todo{to do}
\begin{itemize}
    \item we outline a transformation of COGS logical forms to graphs such that the AM parser can work with them (§\ref{sec:cogs2graph}).
    \item we show how COGS can be solved with a compositional graph-based parser (with little supervision). Whereas BART \cite{lewis-etal-2020-bart-acl}, a pretrained encoder-decoder network, fails to do so despite the big pretraining dataset (§\ref{sec:experiments}). 
    \item by that we demonstrate that the AM parser is indeed compositional and can generalize compositionally.
\end{itemize}

Outline of the paper: 
the compositional parser and the COGS dataset (§\ref{sec:background}), 
the conversion from logical forms to graphs (§\ref{sec:amparsing4cogs}), 
experiments, results and error analysis (§\ref{sec:experiments}), 
discussion (§\ref{sec:discussion}), 
related work (§\ref{sec:related_work}), 
§\ref{sec:conclusion} concludes.
}


\section{Compositional Generalization}\label{sec:background}



Compositional generalization is the ability to determine the meaning
of unseen sentences using compositional principles. Humans can
understand and produce a potentially infinite number of novel
linguistic expressions by dynamically recombining known elements
\citep{chomsky-1957-syntactic,fodor-pylyshyn-1988-connectionism,fodor-lepore-2002-compositionality}.
For semantic parsers, compositional generalization tasks
systematically vary language use between the training and the
generalization set; as such, the system must recombine parts
of multiple training instances to predict the meaning of a single test
instance. 

COGS \cite{kim-linzen-2020-cogs} is a synthetic semantic parsing
dataset in which English sentences must be mapped to logic-based
meaning representations. It distinguishes 21 \emph{generalization
  types}, each of which requires generalizing from training instances
to test instances in a particular systematic and
linguistically-informed way. We follow the authors and distinguish two
classes of generalization types; we further comment on a third class
based on data from model performance.

\begin{table*}[htb!]
    \small
    \centering
    \setlength{\tabcolsep}{2pt}
    \begin{tabularx}{\linewidth}{lXX} \toprule
         & \textbf{Training} & \textbf{Generalization} \\ \midrule
         (a) \lexg
         & {A \underline{hedgehog} ate the cake. \newline
         \lform{*cake($x_4$); \underline{hedgehog($x_1$)} $\land$ eat.agent($x_2, \underline{x_1}$) $\land$ eat.theme($x_2, x_4$)}} 
         & {The baby liked the \underline{hedgehog}. \newline
         \lform{*baby($x_1$); *\underline{hedgehog($x_4$)}; like.agent($x_2, x_1$) $\land$ like.theme($x_2, \underline{x_4}$)} }\\ 
         \midrule
         (b) \propg
         & \underline{Charlie} ate the cake. \newline
         \lform{*cake($x_3$); eat.agent($x_1,$ \underline{Charlie}) $\land$ eat.theme($x_1, x_3$)}
         & The monster ate \underline{Charlie}. \newline
         \lform{*monster($x_1$); eat.agent($x_2, x_1$) $\land$ eat.theme($x_2,$ \underline{Charlie})}\\ 
         \midrule
         (c) \structg
         & Ava saw a ball in a bowl on the table. \newline
         \lform{*table($x_9$); see.agent($x_1,$ Ava) $\land$ see.theme($x_1, x_3$) $\land$ ball($x_3$) $\land$ ball.nmod.in($x_3, x_6$) $\land$ bowl($x_6$) $\land$ bowl.nmod.on($x_6,x_9$)}
         &  Ava saw a ball in a bowl on the table \underline{on the floor}. \newline
         \lform{*table($x_9$); \underline{*floor($x_{12}$);} see.agent($x_1,$ Ava) $\land$ see.theme($x_1, x_3$) $\land$ ball($x_3$) $\land$ ball.nmod.in($x_3, x_6$) $\land$ bowl($x_6$) $\land$ bowl.nmod.on($x_6,x_9$) \underline{$\land$ table.nmod.on($x_9,x_{12}$)}} \\ \cmidrule(l){2-3}
         (d) \structg & Noah ate \underline{the cake on the plate}. \newline
         \lform{*cake($x_3$); *plate($x_6$); \newline eat.agent($x_1,$ Noah) $\land$ eat.theme($x_1, x_3$) $\land$ cake.nmod.on($x_3,x_6$)} 
         & \underline{The cake on the table} burned. \newline
         \lform{*cake($x_1$); *table($x_4$); cake.nmod.on($x_1,x_4$) $\land$ burn.theme($x_3, x_1$)}\\ 
         \bottomrule
    \end{tabularx}
    \caption{
        Some examples from the COGS dataset. Examples (a) represent lexical generalization (\lexg); (b), to object proper noun generalization (\propg); and (c-d), structural generalization (\structg).
    }\label{tab:cogssamples}
\end{table*}

\emph{Lexical generalization} involves recombining known grammatical
structures with words that were not observed in these particular
structures in training.  An example is the generalization type
``subject to object (common)'' (\cref{tab:cogssamples}a), in which a
common noun (``hedgehog'') is only seen as a subject in training,
whereas it is only used as on object in the generalization testset. 
  Note that the syntactic structure at generalization time (e.g.\ that
  of a transitive sentence) was already observed in training. On the
  semantics side, the meaning representations are identical, except
  for replacing some constants and quantifiers and renaming some
  variables. 
  Thus, lexical generalization in COGS amounts to
  learning how to fill fixed templates.

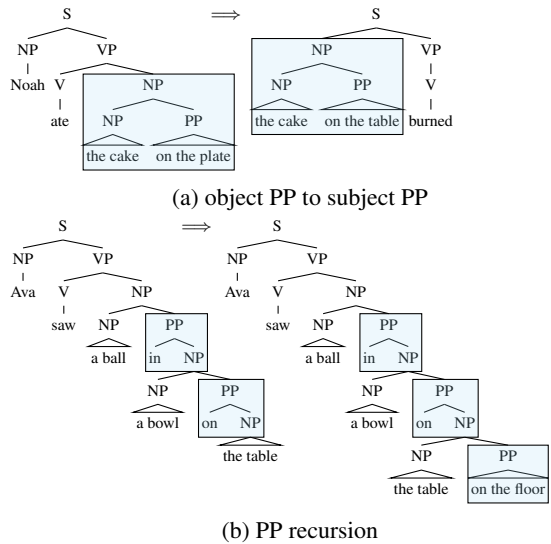
\begin{figure}[t]
  \centering
  \begin{subfigure}[b]{\columnwidth}
    \centering
    \hbox{\tiny

\begin{forest} for tree={l=0, l sep=4, inner sep=1}
[S, s sep=1pt, 
  [NP [Noah]]
  [VP, s sep=10pt, 
    [V [ate]]
    [NP,tikz={\node [draw=black,fill=cyan!30, fill opacity=0.2,inner sep=0,fit to=tree]{};}
      [NP [the cake,roof]]
      [PP [on the plate,roof]]
    ]
  ]
]
\end{forest}
\hspace{-5mm}
$\implies$

\begin{forest} for tree={l=0, l sep=4, inner sep=1}
[S, s sep=1pt,  
  [NP,tikz={\node [draw=black,fill=cyan!30, fill opacity=0.2,inner sep=0,fit to=tree]{};}
      [NP [the cake,roof]]
      [PP [on the table,roof]]
  ]
  [VP
    [V [burned]]
  ]
]
\end{forest}
    }
    \caption{object PP to subject PP}\label{fig:structural-generalization:toSubjPP}
  \end{subfigure}
  
  \begin{subfigure}[b]{\columnwidth}
      \centering
      \hbox{\tiny

\begin{forest} for tree={l=0, l sep=3, s sep=6pt, inner sep=1}
[S 
  [NP [Ava]]
  [VP
    [V [saw]]
    [NP
      [NP [a ball,roof]]
      [PP,tikz={\node [draw=black,fill=cyan!30, fill opacity=0.2,inner sep=0,fit=()(!1)(!2)]{};}
       [in]
       [NP
         [NP [a bowl,roof]]
         [PP,tikz={\node [draw=black,fill=cyan!30, fill opacity=0.2,inner sep=0,fit=()(!1)(!2)]{};}
           [on]
           [NP [the table,roof]]
         ]
       ]
      ]
    ]
  ]
]
\end{forest}
\hspace{-1.5cm}
$\implies$

\begin{forest} for tree={l=0, l sep=3, s sep=6pt, inner sep=1}
[S 
  [NP [Ava]]
  [VP
    [V [saw]]
    [NP
      [NP [a ball,roof]]
      [PP,tikz={\node [draw=black,fill=cyan!30, fill opacity=0.2,inner sep=0,fit=()(!1)(!2)]{};}
       [in]
       [NP
         [NP [a bowl,roof]]
         [PP,tikz={\node [draw=black,fill=cyan!30, fill opacity=0.2,inner sep=0,fit=()(!1)(!2)]{};}
           [on]
           [NP 
             [NP [the table,roof]]
             [PP,tikz={\node [draw=black,fill=cyan!30, fill opacity=0.2,inner sep=0,fit to=tree]{};}
               [on the floor,roof]
             ]
           ]
         ]
       ]
      ]
    ]
  ]
]
\end{forest}
      }
      \caption{PP recursion}\label{fig:structural-generalization:PPrecursion}
  \end{subfigure}
  \caption{Structural generalization in COGS.}
  \label{fig:structural-generalization}
\end{figure}

By contrast, \emph{structural generalization} involves generalizing to
linguistic structures that were not seen in training (cf.\
\cref{tab:cogssamples}c,d). Examples are the generalization types ``PP
recursion'', where training instances contain prepositional phrases of
depth up to two and generalization instances have PPs of depth 3--12;
and ``object PP to subject PP'', where PPs modify only objects in
training and only subjects at test time. These structural changes are
illustrated in 
\cref{fig:structural-generalization}.

  A third class we observe involves
  generalizing to \emph{object usage of proper nouns} (\cref{tab:cogssamples}b). 
  Though technically a subset of lexical generalization, this subgroup is harder than types of the same class (cf.~\cref{subsec:first:experiments}); we report and discuss it separately. 

  Lexical generalization captures a very limited fragment of
  compositionality, in that it only requires to fill a fixed number of
  slots with new values. The key point about compositionality in
  semantics is that language is infinitely productive, and humans can
  assign meaning to new grammatical structures based on finite
  experience. Assigning meaning to unseen structures is exercised only
  by structural types. This distinction is borne out in model
  performance (\cref{subsec:first:experiments}): while lexical
  generalization can be handled by many neural architectures,
  structural generalization requires parsing architectures aware of
  complex sentence structure.


\section{Related Work}\label{sec:related_work}


\citet{kim-linzen-2020-cogs} demonstrate that simple
seq2seq models (LSTMs and Transformers) struggle with all
generalization types in COGS. Subsequent work with novel seq2seq
architectures achieve a much higher mean accuracy on the COGS
generalization set \citep{akyurek-andreas-2021-lexicon,csordas-etal-2021-devil,
  conklin-etal-2021-meta,tay-etal-2021-pretrained,orhan-2021-compgen,
  zheng-lapata-2021-disentangled}, but their accuracy on the
generalization set still lags more than ten points behind that on the
in-domain test set.


COGS can also be addressed with \emph{compositional models}, which
directly model linguistic structure and implement the Principle of
Compositionality. The LeAR model of \citet{liu-etal-2021-learning}
achieves a generalization accuracy of 98\%, outperforming all known
seq2seq models by at least ten points. LeAR also sets new states of
the art on CFQ and Geoquery, but has not been demonstrated to be
applicable to broad-coverage semantic parsing.

Compositional semantic parsers for other tasks include the AM parser
\citep{groschwitz-etal-2018-amr,lindemann-etal-2020-fast}
(\cref{sec:amparsing4cogs}) and SpanBasedSP
\citep{herzig-berant-2021-span-based}.  The AM parser has been shown
to achieve high accuracy and parsing speed on broad-coverage semantic
parsing datasets such as the AMRBank.  SpanBasedSP parses Geoquery,
SCAN, and CLOSURE accurately through unsupervised training of a
span-based chart parser. \citet{shaw-etal-2021-compositional} combine
quasi-synchronous context-free grammars with the T5 language model to
obtain even higher accuracies on Geoquery, demonstrating some
generalization from easy training examples to hard test instances.

Structural generalization has also been probed in syntactic parsing tasks.
\citet{linzen2016assessing} define
a number-prediction task that requires learning syntactic
structure and find that LSTMs perform with some success;
however, \citet{kuncoro-etal-2018-lstms} find that structure-aware
RNNGs perform this task more accurately.
\citet{DBLP:journals/tacl/McCoyFL20} found that hierarchical representations are necessary for human-like syntactic generalizations on a question formation
task, which seq2seq models cannot learn.




\section{Parsing COGS with the AM parser}\label{sec:amparsing4cogs}


\subsection{The AM parser}

To better understand how compositional models perform on compositional generalization, we adapt the broad-coverage AM parser to COGS.
The AM parser \cite{groschwitz-etal-2018-amr} is a compositional
semantic parser that learns to map sentences to graphs. It was the
first semantic parser to perform with high accuracy across all major
graphbanks \citep{lindemann-etal-2019-compositional} and can achieve
very high parsing speeds \citep{lindemann-etal-2020-fast}. Thus, though not yet tested on synthetic generalization sets, the AM parser exhibits the ability to handle natural language and related generalizations in the wild.

\begin{figure}
  \centering
  \includegraphics[width=\columnwidth]{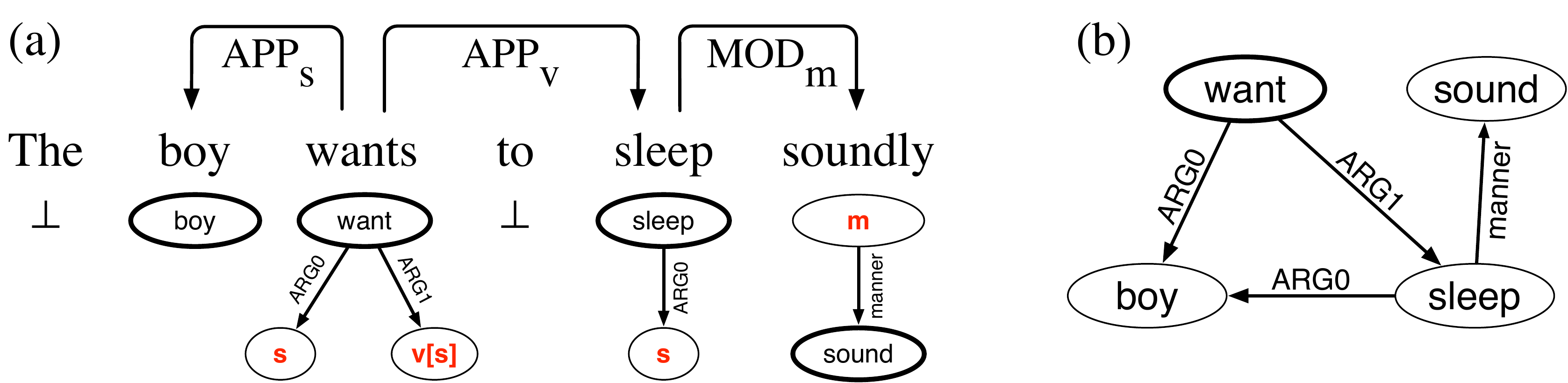}
  \caption{(a) AM dependency tree with (b) its value.}
  \label{fig:am}
\end{figure}

Instead of predicting the graph directly, the AM parser first predicts
a graph fragment for each token in the sentence and a (semantic)
dependency tree that connects them. This is illustrated in
\cref{fig:am}a; words that do not contribute to the sentence
meaning are tagged with $\bot$. This dependency tree is then
evaluated deterministically into a graph (\cref{fig:am}b) using
the operations of the \emph{AM algebra}. The ``Apply'' operation fills
an argument slot of a graph (drawn in red) by inserting the root node (drawn with a bold outline)
of another graph into this slot; for instance, this is how the \app{s}
operation inserts the ``boy'' node into the ARG0 of ``want''. The
``Modify'' operation attaches a modifier to a node; this is how the
\modify{m} operation attaches the ``manner-sound'' graph to the ``sleep''
node. The dependency tree captures how the meaning of the
sentence can be compositionally obtained from the meanings of the
words.

AM parsing is done by combining a neural dependency
parser with a neural tagger for predicting the graph fragments. We follow \citet{lindemann-etal-2019-compositional} and rely on the
dependency parsing model of \citet{kiperwasser-goldberg-2016-simple},
which scores each dependency edge by feeding neural representations for the
two tokens to an MLP.
We follow the setup of \citet{groschwitz-etal-2021-learning}, which
does not require explicit annotations with AM dependency trees, to
train the parser.

\subsection{AM parsing for  COGS}\label{subsec:cogs2graph}
\begin{figure}[t] 
    \centering
    \lform{\scriptsize
    *table($x_9$); see.agent($x_1,$ Ava) $\land$ see.theme($x_1, x_3$) $\land$ ball($x_3$) $\land$ ball.nmod.in($x_3, x_6$) $\land$ bowl($x_6$) $\land$ bowl.nmod.on($x_6,x_9$)
    }\\[8pt]
    \includegraphics[width=0.9\columnwidth]{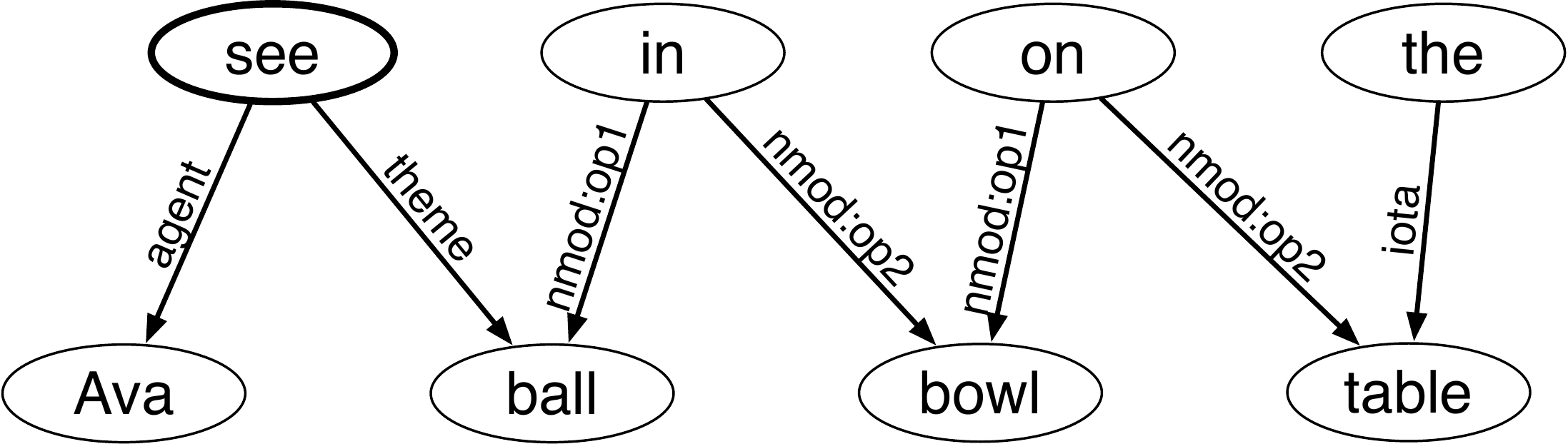}
    \caption{
        Logical form to graph conversion for \sentence{Ava saw a ball in a bowl on the table} (cf.~\cref{tab:cogssamples}c).
    }\label{fig:cogsgraph}
\end{figure}

We apply the AM parser to COGS by converting the semantic
representations in COGS to graphs. The conversion is illustrated in
\cref{fig:cogsgraph}.

Given a logical form of COGS, we create a graph that has one node for
each variable $x_i$ and each constant (e.g. \lform{Ava}). If a
variable appears as the first argument of an atom of the
form $\lform{pred}.\lform{arg}(x,y)$, we assign it the node label
$\lform{pred}$ in the graph. We also add an edge from $x$ to $y$ with
label $\lform{arg}$. 
E.g.~\lform{see.agent($x_1,$ Ava)} turns into an \graphedge{agent} edge from \graphnode{see} to \graphnode{Ava}.
Each \emph{iota term}
\lform{*noun($x_{\mathrm{noun}}$)} is treated as an edge from a fresh
node with label ``the'' to $x_{\text{noun}}$. Preposition meaning
\lform{bowl.nmod.on($x_6,x_9$)}
is
represented as a node (labeled \graphnode{on}) with outgoing edges to the
two arguments/nouns (\graphedge{nmod.op1} to \word{bowl},
\graphedge{nmod.op2} to \word{table}). 




By encoding the logical form as a graph, we lose the ordering of the
conjuncts. The \sortof{correct} order is restored in postprocessing.
More details and graph conversion examples are in
\cref{sec:details_cogs2graph}.

\section{Experiments on COGS}\label{subsec:first:experiments}

With two compositional models available on COGS, we can now compare
compositional semantic parsers, as a class, to seq2seq models, as
a class, on compositional generalization in COGS.





\subsection{Experimental setup}\label{subsec:first:setup}

We follow standard COGS practice and evaluate all models on both the
(in-distribution) test set and the generalization set. In addition to
the regular COGS training set (\sortof{train}) of 24,155 training
instances, we also report numbers for models trained on the extended
training set \sortof{train100}, of 39,500 instances
\citep[Appendix E.2]{kim-linzen-2020-cogs}.  The \sortof{train100}
set extends \sortof{train} with 100 copies of each exposure
example. For instance, for the generalization instance in
Table~\ref{tab:cogssamples}a, \sortof{train100} will contain 100 different
sentences in which ``the/a hedgehog'' appears as subject (rather than
just one in `train'). 
We report exact match accuracies, averaged across 5 training runs,
along with their standard deviations.

{
\begin{table}[t]
\footnotesize 
    \centering
    \setlength{\tabcolsep}{2pt}
    \begin{tabular}{@{}llrrrr@{}} \toprule
      && \multicolumn{2}{c}{train} & \multicolumn{2}{c}{train100} \\
      \cmidrule(lr){3-4} \cmidrule(lr){5-6}
      && Test & Gen & Test & Gen \\ \midrule
      \parbox[t]{2mm}{\multirow{7}{*}{\rotatebox[origin=c]{90}{seq-to-seq}}}
      &\citealt{kim-linzen-2020-cogs} & \oldNumber{96} & \oldNumber{35} & \oldNumber{94} & \oldNumber{63} \\
      &\citealt{conklin-etal-2021-meta}  & \oldNumber{99} & \oldNumber{66.7} & 99 & 75.4 \\
      &\citealt{csordas-etal-2021-devil} & \oldNumber{100} & \oldNumber{81} & - & 75.4 \\
      &\citealt{akyurek-andreas-2021-lexicon} & - & \oldNumber{83} & 99 & 84.5 \\
      &\citealt{zheng-lapata-2021-disentangled} $^\dagger$ & - & \oldNumber{87.9} & - & - \\ 
      &\citealt{orhan-2021-compgen} $^\dagger$ & - & \oldNumber{84.6} & - & - \\
      &\citealt{tay-etal-2021-pretrained}  $^\dagger$ & \oldNumber{95} & \oldNumber{77.5} & - & -
      \\
      \cmidrule(l){2-6}
      &BART  $^\dagger$ & 100 & 77.5\stderr{0.4}  & 100 & 82.7\stderr{1.4} \\
      &BART+syn  $^\dagger$ & 100 & 80.2\stderr{0.4}  & 100 & 85.9\stderr{0.3} \\
      \midrule
      \parbox[t]{2mm}{\multirow{5}{*}{\rotatebox[origin=c]{90}{compositional}}}      
      &\citealt{liu-etal-2021-learning}: LeAR\protect\footnotemark & - & \textbf{98.9}\stderr{0.9} & - & - \\
      \cmidrule(l){2-6}
      &\amToken & 100 & 59.9\stderr{\hspace{.5em}2.7} & 100 & 91.1\stderr{2.3} \\ 
      &\amToken+\dist & 100 & 62.6\stderr{10.8} & 100 & 88.6\stderr{4.9}\\
      &\amBert\  $^\dagger$ & 100 & 79.6\stderr{\hspace{.5em}6.4} & 100 & 93.6\stderr{1.4} \\
      &\amBert+\dist  $^\dagger$ & 100 & 78.3\stderr{22.9} & 100 & \textbf{98.4}\stderr{0.9} \\
      \bottomrule
    \end{tabular}
    \caption{
      Exact match accuracies on COGS. Results in gray are taken from the respective papers. $^\dagger$) models that use pretraining. 
    }\label{tab:results}
  \end{table}
}

\begin{table*}
      \centering 
      \scriptsize 
      \setlength{\tabcolsep}{2pt}
      
\begin{tabular}{@{}lll|ccc|cc|c|c@{}}
\toprule
& & Class & \multicolumn{3}{c|}{\structg} & \multicolumn{2}{c|}{\propg} & \lexg & \\
& & Gen. type & Obj to Subj PP & CP recursion & PP recursion & prim to
                                                               obj
                                                               (proper)
  & subj to obj (proper) & all 16 other types & Overall \\ 
\midrule
\parbox[t]{2mm}{\multirow{12}{*}{\rotatebox[origin=c]{90}{semantics}}} 
& \amBert & train100 & \asbar{49} & \asbar{100} & \asbar{41} & \asbar{85} & \asbar{90} & \asbar{100} & \asbar{94} \\
& \amBert+\dist & train100 & \asbar{78} & \asbar{100} & \asbar{99} & \asbar{94} & \asbar{96} & \asbar{100} & \asbar{98} \\
& LeAR & train & \asbar{93} & \asbar{100} & \asbar{99} & \asbar{93} & \asbar{93} & \asbar{100} & \asbar{99} \\ 
\cmidrule{2-10}
& \citealt{kim-linzen-2020-cogs} & train & \asbar{0} & \asbar{0} & \asbar{0} & \asbar{0} & \asbar{30} & \asbar{45} & \asbar{35} \\
& \citealt{akyurek-andreas-2021-lexicon} & train & \asbar{0} & \asbar{0} & \asbar{1} & \asbar{66} & \asbar{64} & \asbar{100} & \asbar{82} \\
& \citealt{orhan-2021-compgen} & train & \asbar{0} & \asbar{0} & \asbar{10} & \asbar{84} & \asbar{86} & \asbar{100} & \asbar{85} \\
& \citealt{zheng-lapata-2021-disentangled} & train & \asbar{0} & \asbar{12} & \asbar{39} & \asbar{92} & \asbar{91} & \asbar{100} & \asbar{89} \\
& \citealt{kim-linzen-2020-cogs} & train100 & \asbar{0} & \asbar{0} & \asbar{0} & \asbar{23} & \asbar{54} & \asbar{77} & \asbar{63} \\
& \citealt{conklin-etal-2021-meta} & train100 & \asbar{0} & \asbar{0} & \asbar{0} & \asbar{20} & \asbar{66} & \asbar{94} & \asbar{75} \\
& \citealt{csordas-etal-2021-devil} & train100 & \asbar{0} & \asbar{0} & \asbar{0} & \asbar{55} & \asbar{62} & \asbar{92} & \asbar{75} \\
& BART & train100 & \asbar{0} & \asbar{0} & \asbar{10} & \asbar{55} & \asbar{86} & \asbar{99} & \asbar{83} \\
& BART+syn & train100 & \asbar{0} & \asbar{7} & \asbar{8} & \asbar{98} & \asbar{95} & \asbar{100} & \asbar{86} \\ 
\midrule
\parbox[t]{2mm}{\multirow{2}{*}{\rotatebox[origin=c]{90}{syntax}}} & Benepar & train100 & \asbar{84} & \asbar{95} & \asbar{98} & \asbar{99} & \asbar{100} & \asbar{100} & \asbar{99} \\
\cmidrule{2-10}
& BART & train100 & \asbar{1} & \asbar{4} & \asbar{8} & \asbar{97} & \asbar{94} & \asbar{96} & \asbar{83} \\
\bottomrule
\end{tabular} 
    \caption{Exact match accuracies on the individual generalization types. We have compressed all 16 generalization types of the \lexg class into a single column and report the average accuracy.
    }\label{tab:selected_gentype_eval}
  \end{table*}


\paragraph{Sequence-to-sequence models.} \label{subsec:seq2seq:setup}
We train BART \cite{lewis-etal-2020-bart-acl} as a semantic parser on
COGS. This is a strong representative of the family of seq2seq models,
as a slightly extended form of BART \cite{bevilacqua-etal-2021-one}
set a new state of the art on
semantic
parsing on the AMR corpus \cite{banarescu-etal-2013-abstract}. 
To apply BART on COGS, we directly fine-tune the pretrained
\textit{bart-base} model on it with the corresponding tokenizer. Training details are described in \cref{sec:training_details_bart}.


We also report results for all other published seq2seq models for COGS
\citep{kim-linzen-2020-cogs,conklin-etal-2021-meta,csordas-etal-2021-devil,akyurek-andreas-2021-lexicon,tay-etal-2021-pretrained,orhan-2021-compgen,zheng-lapata-2021-disentangled}.
We retrained some of these models on train100 to measure the impact of
the training set.


\paragraph{Compositional models.}
We train the AM parser on the COGS graph corpus
(cf.~\cref{subsec:cogs2graph}) and copied most hyperparameter values
from \citet{groschwitz-etal-2021-learning}'s  training setup for AMR to make overfitting to COGS less likely; details are described in
\cref{sec:training_details_amparser}. 

The AM parser either receives pretrained word embeddings from BERT
\citep{devlin-etal-2019-bert} (\sortof{\amBert}) or learns embeddings
from the COGS data only (\sortof{\amToken}). We run the training
algorithm with up to three argument slots to enable the analysis of
ditransitive verbs. For evaluation, we revert the graph conversion to
reconstruct the logical forms.
  \footnotetext{All LeAR numbers are based on our reproduction of
    their COGS evaluation; they report an accuracy of 97.7.}

For PP recursion, COGS
eliminates potential PP attachment ambiguities and
assumes that each PP modifies the noun immediately to its left.  
We hypothesize that explicit distance information between tokens could
help the AM parser learn this regularity: 
Instead of passing only the representations of the
potential parent and child node to the edge-scoring model, we also
pass an encoding of their relative distance in the string
\cite{vaswani-etal-2017-attention}, yielding the AM parser models with the ``+\dist'' suffix. 

Finally, we report evaluation results for LeAR, the compositional COGS parser of
\citet{liu-etal-2021-learning}.

\subsection{Results}\label{subsec:results}

The results are summarized in \Cref{tab:results}.

\paragraph{Compositional outperforms seq2seq.}
While all models achieve near-perfect accuracy on the in-distribution
test sets, we find that when trained on \sortof{train100}, all
compositional models outperform all seq2seq models on the
generalization set, by a wide margin. This includes the very strong
BART baseline, which holds the state of the art in broad-coverage
parsing for AMR.

LeAR even achieves its near-perfect accuracy when trained on `train',
and outperforms all seq2seq models trained on either dataset. See
below for a detailed discussion of the AM parser.


\paragraph{Performance by generalization type.}
To understand this result more clearly, we break down the accuracy by
generalization type. This analysis is shown in
\cref{tab:selected_gentype_eval}.  We will explain ``BART+syn'' in
\cref{subsec:seq2seq:semantic_gen_with_syntax} and the ``syntax''
rows in \cref{subsec:seq2seq:syntactic_gen}. We compare the
compositional models against all seq2seq models that report these
fine-grained numbers or for which they were easy to reproduce (see
\cref{sec:training_details_bart,sec:detailed_evaluation_results} for details).

The results group neatly with the three classes of generalization types
outlined in \cref{sec:background}: \textsc{Lex}, \textsc{Struct}, and \textsc{Prop}. All recent models achieve
near-perfect accuracy on each of the 16 lexical generalization
types. On structural generalization types, seq2seq models achieve very
low accuracies, whereas the compositional parsers (\amBert+\dist and LeAR)
are still very accurate. The proper-noun object cases are somewhere in
the middle, with the seq2seq models reporting middling numbers.

\begin{figure}
    \centering
    \includegraphics[scale=.4,trim={5mm 5mm 40mm 0},clip]{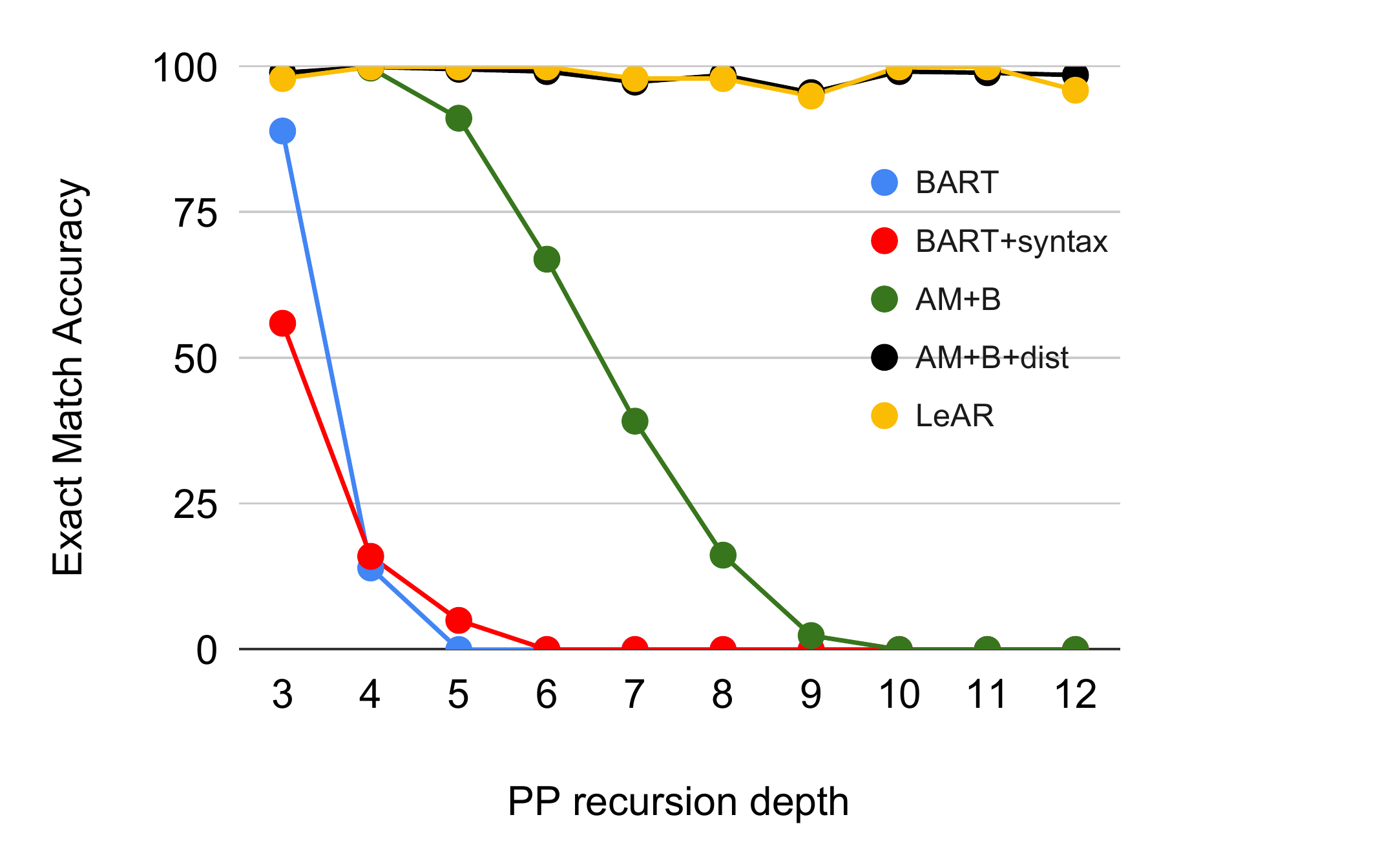}
    \caption{Influence of PP recursion depth on overall PP depth
      generalization accuracy.
    }\label{fig:depths_plot}
  \end{figure}

\paragraph{Depth generalization (recursion).}
There is a particularly pronounced difference between compositional
and seq2seq models on the two ``recursion'' generalization types (cf.\
\cref{fig:structural-generalization:PPrecursion}). In these cases,
the training data contains examples up to depth two and the
generalization data has depths 3--12. \Cref{fig:depths_plot} shows
the accuracy of several models on PP recursion in detail. As we see,
the accuracy of BART (even when informed by syntax, cf.\
\cref{subsec:seq2seq:semantic_gen_with_syntax}) degrades
quickly with recursion depth. By contrast, both LeAR and \amBert+\dist
maintain their high accuracy across all recursion depths. This
suggests that they learn the correct structural generalizations even
from training observations of limited depth.

\paragraph{Effect of distance encoding for AM parser.}
As illustrated in \cref{fig:depths_plot}, the accuracy of the
unmodified AM parser without the distance feature degrades with
increasing PP recursion depth. An error analysis showed that this is
because the AM parser is uncertain about the attachment of PPs in the
middle of the string, confirming our hypothesis that it does not learn
the idiosyncratic treatment of PPs in COGS (always attach
low). Adding the distance feature solves this problem.

There is an interesting asymmetry between the behavior of the AM
parser on PP recursion and CP recursion, which nests sentential
complements within each other (``Emma said that Noah knew that the cat
danced''): The accuracy of the unmodified AM parser is stable across
recursion depths for CP recursion, and the distance feature is only
needed for PPs. This can be explained by the way in which the AM
parser learns to incorporate PPs and CPs into the dependency tree: it
uses \app{} edges to combine verbs with CPs, which ensures that only a
single CP can be combined with each sentence-embedding verb. By
contrast, each NP can be modified by an arbitrary number of PPs using
\modify{} edges. Thus a confusion over attachment is only possible for
PPs, not CPs.

\paragraph{Effect of training regime.}
Parsers on COGS are traditionally not allowed any pretraining
\cite{kim-linzen-2020-cogs}, in order to judge their ability to
generalize from  limited observations. We see in the
experiments above that the use of pretrained word embeddings
helps the AM parser achieve accuracy parity with LeAR, but is not
needed to outperform all seq2seq models on `train100'.

Training on \sortof{train100} helps the AM parser more than any other
model in \cref{tab:results}. The difference between its accuracy on
`train' and `train100' is due to lexical issues: we found that when
trained on `train', the AM parser typically predicts the correct
delexicalized formulas and then inserts an incorrect but related
constant or predicate symbol (e.g.\ \word{Emma} instead of
\word{Charlie} in Table~\ref{tab:cogssamples}b). Trained on `train',
\amBert+\dist achieves a mean accuracy on \structg\ of 89.6 (compared
to 92.3 for `train100'), whereas the mean accuracy on \lexg\ drops to
76. Even without BERT and trained on `train', \amToken+\dist gets 74.6 on
\structg, drastically outperforming the seq2seq models
(Appendix~\ref{sec:detailed_evaluation_results}).

\section{The role of syntax}
\label{sec:analysis-extensions}

Our finding that seq2seq models perform so poorly on structural
generalization in COGS begs the question: Is there anything special
about the meaning representations in COGS that makes structural
generalization hard, or would seq2seq models struggle similarly on
other target representations for these generalization types? Do
seq2seq models have a specific weakness regarding semantic
compositionality? Or is it because they systematically lack a bias
that would help them generalize over structure in language? In this
section, we investigate these questions by recasting COGS as a
syntactic corpus.



\subsection{Syntactic  generalization}\label{subsec:seq2seq:syntactic_gen}

We obtain a syntactic annotation for each instance in COGS from the
(unambiguous) original PCFG grammar used to generate COGS (cf.\
\cref{fig:structural-generalization}).  We replace the very
fine-grained non-terminals (e.g.\  \lform{NP\_animate\_dobj\_noPP}) of
the original PCFG with more general ones (e.g.\ \lform{NP}) and remove
duplicate rules (e.g.\ \lform{NP}$\to$\lform{NP}) resulting from this.
We train BART on predicting linearized constituency trees from the
input strings. For comparison, we also train the Neural Berkeley
Parser \citep{kitaev-klein-2018-constituency} on COGS syntax
(``Benepar'' in the tables). This parser consists of a self-attention
encoder and a chart decoder. It is therefore \emph{structure-aware},
in that it explicitly models tree structures; this is the analogue of
a compositional parser for semantics.

Results are shown in the two bottom rows of
\cref{tab:selected_gentype_eval}. We find the same pattern as in the
semantic parsing case: the seq2seq model does well on \propg and
\lexg, but struggles with \structg. The structure-aware Berkeley
parser handles all three generalization types well. Thus, the
difficulties that seq2seq models have on structural generalization on
COGS are not limited to semantics: rather, they seem to be a general
limitation in the ability of seq2seq models to learn linguistic
structure from structurally simple examples and use it
productively. Not only does compositional generalization require
compositional parsers; structural generalization in semantics or
syntax seems to require parsers which are aware of that structure.


\subsection{Compositional generalization from correct
  syntax}\label{subsec:seq2seq:semantic_gen_with_syntax}

But perhaps the poor performance of seq2seq semantic parsers on
\structg\ is caused \emph{only} by their inability to learn to
generalize syntactically? Would their accuracy catch up with that of
compositional models if we gave them access to syntax?

We retrained BART on predicting semantic representations, but
instead of feeding it the raw sentence, we provide as input the
linearized gold constituency tree (``\lform{(NP ({Det} a) ({N} rose))}''),
both for training and inference.
 This method is similar to
\citet{li-etal-2017-modeling} and \citet{currey-heafield-2019-incorporating},
but we allow attention over special tokens such as ``\lform{(}'' during decoding.

We report the results as ``BART+syn'' in \cref{tab:results} and
\cref{tab:selected_gentype_eval}; the overall accuracy increases by
3.2\% over BART. This
is mostly because providing the syntax tree allows BART to generalize
correctly on \propg. 
However, \structg remains out
of reach for BART+syn, confirming the deep difficulty of
structural generalization for seq2seq models.

We also explored other ways to inform BART with syntax, through
multi-task learning \cite{sennrich-etal-2016-controlling,
  currey-heafield-2019-incorporating} and syntax-based masking in the
self-attention encoder \cite{kim-etal-2021-improving}. Neither method
substantially improved the accuracy of BART on the COGS generalization
set (+1.4\% and +2.1\% overall accuracy, respectively).  More
detailed results are in \cref{sec:detailed_evaluation_results}.




\section{Discussion}\label{sec:discussion}

\paragraph{Compositional generalization requires compositional parsers.}
\Cref{tab:selected_gentype_eval} paints a clear picture: compositional
generalization in COGS can be solved by semantic parsers that have
compositionality built in, but seq2seq models perform poorly on
structural generalization. This remains true even for seq2seq models
that are known to perform well on semantic parsing, for syntactic
rather than semantic generalization, and for seq2seq models that are
biased towards learning structure-aware representations by
incorporating information about syntax. Obviously, statements about
entire classes of models must be made with care. But when despite the
best efforts of an active research community \emph{all} seq2seq models
underperform the compositional models, that seems like rather strong
evidence.

Our results are surprising, in that seq2seq models have been shown
through probing tasks to learn some linguistic structure, both with
respect to syntax \cite{blevins-etal-2018-deep} and semantics
\cite{tenney-etal-2019-bert}. At the same time, as mentioned above,
seq2seq models like BART perform very well on broad-coverage tasks
such as AMR parsing. It is an interesting question for future research
to reconcile the ability of seq2seq models to learn soft structural
information with their apparent difficulties in exploiting this
ability to generalize structurally; perhaps their ability to learn
structure rests on the variety of structures observed in
broad-coverage training sets, but not in COGS.

\paragraph{Focus on structural generalization.}
Our experiments indicate that \structg is consistently harder than
\propg and \lexg with respect to generalization accuracy. Not only
is \lexg\ essentially a solved problem; but as we discussed in
Section~\ref{sec:background}, the infinitely productive nature of full
compositionality is only captured by structural types of
generalization. Compositionality is not just about using new and
similar words in known structures (slot filling), but also about
building new, acceptable structures based on known ones.

When papers only report the mean accuracy of a system across all
generalization types, the accuracy on the 16 lexical generalization
types overshadows the accuracy on the three structural generalization
types. The overall accuracy can make systems look more capable of
compositional generalization than they really are.

Future work on compositional generalization will benefit from (i)
reporting the accuracy on structural generalization tasks separately
and (ii) expanding datasets that test compositional generalization to
include more types of structural generalization. We hope to offer such
a dataset in future research.

\paragraph{What's so difficult about objPP to subjPP?}
``ObjPP to subjPP'' is the most challenging generalization type across
all models. It is illuminating to investigate the errors that happen
here, as they differ across models.

Table~\ref{tab:erranalysis} shows typical errors of BART and the AM
parser. The AM parser chooses to use the most recent simple NP (``the
house'') as the agent of ``scream'' and then attaches ``the baby on a
tray'' in some random place. By contrast, BART analyzes the sentence
as ``the baby screamed the tray on the house'', preferring to reuse the pattern for object-PP sentences even if the
intransitive verb does not license it. BART also displays an unawareness
of word order that is reminiscent of the difficulties that seq2seq
models otherwise face in relating syntax to word order
\citep{DBLP:journals/tacl/McCoyFL20}.

We see from both examples that ``objPP to\linebreak subjPP'' involves major
structural changes to the formula that must be grounded in both
lexical (verb valency) and structural (word order)
information. Developing a model that learns to do this with perfect
accuracy remains an interesting challenge.


\begin{table}[t]
\centering
\footnotesize
\begin{tabularx}{\linewidth}{lX} \toprule
Gold &  \lform{*baby($x_1$); *house($x_7$); baby.nmod.on($x_1, x_4$) $\land$ tray($x_4$) $\land$ tray.nmod.in($x_4, x_7$) $\land$ scream.agent($x_8, x_1$)} \\
BART & \lform{*baby($x_1$); *house($x_7$); scream.agent($\textcolor{red}{x_2}, x_1$) $\land$ scream.\textcolor{red}{theme}($\textcolor{red}{x_2}, x_4$) $\land$ tray($x_4$) $\land$ tray.nmod.in($x_4, x_7$)} \\
AM & \lform{*baby($x_1$); *house($x_7$); baby.nmod.on($x_1, x_4$) $\land$ tray($x_4$) $\land$ tray.nmod.in($x_4, \textcolor{red}{x_8}$) $\land$ scream.agent($x_8 , \textcolor{red}{x_7}$)} \\ \bottomrule
\end{tabularx}
\caption{Error analysis for the sentence \sentence{The baby on a tray in the house screamed}.}\label{tab:erranalysis}
\end{table}

\section{Conclusion}\label{sec:conclusion}

We have shown that compositional semantic parsers systematically
outperform recent seq2seq models on structural generalization in
COGS. While both BART and the AM parser support accurate
broad-coverage semantic parsing, we find that BART struggles with
structural compositional generalization as much as other seq2seq
models, whereas the compositional AM parser
achieves state-of-the-art generalization accuracy on COGS.

These results suggests that even powerful seq2seq
models lack a structural bias that is required to generalize across
linguistic structures as humans do. This lack of bias is not limited
to semantics; our findings indicate that seq2seq models struggle just
as hard to learn syntactic generalizations that are easy for
structure-aware models. Given that all recent models are accurate on
most generalization types, we suggest focusing future evaluations on a
model's accuracy on structural generalization types, and perhaps
extend COGS to a corpus that offers a greater variety of these.






\bibliography{compgen}

\begin{thebibliography}{37}
\expandafter\ifx\csname natexlab\endcsname\relax\def\natexlab#1{#1}\fi

\bibitem[{Akyürek and Andreas(2021)}]{akyurek-andreas-2021-lexicon}
Ekin Akyürek and Jacob Andreas. 2021.
\newblock \href {https://aclanthology.org/2021.acl-long.382} {Lexicon learning
  for few shot sequence modeling}.
\newblock In \emph{Proceedings of the 59th Annual Meeting of the Association
  for Computational Linguistics and the 11th International Joint Conference on
  Natural Language Processing (Volume 1: Long Papers)}, pages 4934--4946,
  Online. Association for Computational Linguistics.

\bibitem[{Banarescu et~al.(2013)Banarescu, Bonial, Cai, Georgescu, Griffitt,
  Hermjakob, Knight, Koehn, Palmer, and
  Schneider}]{banarescu-etal-2013-abstract}
Laura Banarescu, Claire Bonial, Shu Cai, Madalina Georgescu, Kira Griffitt, Ulf
  Hermjakob, Kevin Knight, Philipp Koehn, Martha Palmer, and Nathan Schneider.
  2013.
\newblock \href {https://www.aclweb.org/anthology/W13-2322} {{A}bstract
  {M}eaning {R}epresentation for sembanking}.
\newblock In \emph{Proceedings of the 7th Linguistic Annotation Workshop and
  Interoperability with Discourse}, pages 178--186. Association for
  Computational Linguistics.

\bibitem[{Bevilacqua et~al.(2021)Bevilacqua, Blloshmi, and
  Navigli}]{bevilacqua-etal-2021-one}
Michele Bevilacqua, Rexhina Blloshmi, and Roberto Navigli. 2021.
\newblock \href {https://ojs.aaai.org/index.php/AAAI/article/view/17489} {One
  {SPRING} to rule them both: {S}ymmetric {AMR} semantic parsing and generation
  without a complex pipeline}.
\newblock In \emph{Proceedings of the AAAI Conference on Artificial
  Intelligence (AAAI-21)}, volume~35, pages 12564--12573. AAAI Press.

\bibitem[{Blevins et~al.(2018)Blevins, Levy, and
  Zettlemoyer}]{blevins-etal-2018-deep}
Terra Blevins, Omer Levy, and Luke Zettlemoyer. 2018.
\newblock \href {https://doi.org/10.18653/v1/P18-2003} {Deep {RNN}s encode soft
  hierarchical syntax}.
\newblock In \emph{Proceedings of the 56th Annual Meeting of the Association
  for Computational Linguistics (Volume 2: Short Papers)}, pages 14--19,
  Melbourne, Australia. Association for Computational Linguistics.

\bibitem[{Chomsky(1957)}]{chomsky-1957-syntactic}
Noam Chomsky. 1957.
\newblock \href {https://doi.org/10.1515/9783112316009} {\emph{Syntactic
  Structures}}.
\newblock De Gruyter Mouton.

\bibitem[{Conklin et~al.(2021)Conklin, Wang, Smith, and
  Titov}]{conklin-etal-2021-meta}
Henry Conklin, Bailin Wang, Kenny Smith, and Ivan Titov. 2021.
\newblock \href {https://aclanthology.org/2021.acl-long.258} {Meta-learning to
  compositionally generalize}.
\newblock In \emph{Proceedings of the 59th Annual Meeting of the Association
  for Computational Linguistics and the 11th International Joint Conference on
  Natural Language Processing (Volume 1: Long Papers)}, pages 3322--3335,
  Online. Association for Computational Linguistics.

\bibitem[{Csord{\'a}s et~al.(2021)Csord{\'a}s, Irie, and
  Schmidhuber}]{csordas-etal-2021-devil}
R{\'o}bert Csord{\'a}s, Kazuki Irie, and Juergen Schmidhuber. 2021.
\newblock \href {https://aclanthology.org/2021.emnlp-main.49} {The devil is in
  the detail: Simple tricks improve systematic generalization of transformers}.
\newblock In \emph{Proceedings of the 2021 Conference on Empirical Methods in
  Natural Language Processing}, pages 619--634, Online and Punta Cana,
  Dominican Republic. Association for Computational Linguistics.

\bibitem[{Currey and Heafield(2019)}]{currey-heafield-2019-incorporating}
Anna Currey and Kenneth Heafield. 2019.
\newblock \href {https://doi.org/10.18653/v1/W19-5203} {Incorporating source
  syntax into transformer-based neural machine translation}.
\newblock In \emph{Proceedings of the Fourth Conference on Machine Translation
  (Volume 1: Research Papers)}, pages 24--33, Florence, Italy. Association for
  Computational Linguistics.

\bibitem[{Devlin et~al.(2019)Devlin, Chang, Lee, and
  Toutanova}]{devlin-etal-2019-bert}
Jacob Devlin, Ming-Wei Chang, Kenton Lee, and Kristina Toutanova. 2019.
\newblock \href {https://doi.org/10.18653/v1/N19-1423} {{BERT}: Pre-training of
  deep bidirectional transformers for language understanding}.
\newblock In \emph{Proceedings of the 2019 Conference of the North {A}merican
  Chapter of the Association for Computational Linguistics: Human Language
  Technologies, Volume 1 (Long and Short Papers)}, pages 4171--4186,
  Minneapolis, Minnesota. Association for Computational Linguistics.

\bibitem[{Fodor and Lepore(2002)}]{fodor-lepore-2002-compositionality}
Jerry~A. Fodor and Ernest Lepore. 2002.
\newblock \emph{The Compositionality Papers}.
\newblock Oxford University Press.

\bibitem[{Fodor and Pylyshyn(1988)}]{fodor-pylyshyn-1988-connectionism}
Jerry~A. Fodor and Zenon~W. Pylyshyn. 1988.
\newblock \href {https://doi.org/10.1016/0010-0277(88)90031-5} {Connectionism
  and cognitive architecture: A critical analysis}.
\newblock \emph{Cognition}, 28(1):3--71.

\bibitem[{Groschwitz et~al.(2021)Groschwitz, Fowlie, and
  Koller}]{groschwitz-etal-2021-learning}
Jonas Groschwitz, Meaghan Fowlie, and Alexander Koller. 2021.
\newblock \href {https://doi.org/10.18653/v1/2021.spnlp-1.3} {Learning
  compositional structures for semantic graph parsing}.
\newblock In \emph{Proceedings of the 5th Workshop on Structured Prediction for
  NLP (SPNLP 2021)}, pages 22--36, Online. Association for Computational
  Linguistics.

\bibitem[{Groschwitz et~al.(2018)Groschwitz, Lindemann, Fowlie, Johnson, and
  Koller}]{groschwitz-etal-2018-amr}
Jonas Groschwitz, Matthias Lindemann, Meaghan Fowlie, Mark Johnson, and
  Alexander Koller. 2018.
\newblock \href {https://doi.org/10.18653/v1/P18-1170} {{AMR} dependency
  parsing with a typed semantic algebra}.
\newblock In \emph{Proceedings of the 56th Annual Meeting of the Association
  for Computational Linguistics (Volume 1: Long Papers)}, pages 1831--1841,
  Melbourne, Australia. Association for Computational Linguistics.

\bibitem[{Herzig and Berant(2021)}]{herzig-berant-2021-span-based}
Jonathan Herzig and Jonathan Berant. 2021.
\newblock \href {https://aclanthology.org/2021.acl-long.74} {Span-based
  semantic parsing for compositional generalization}.
\newblock In \emph{Proceedings of the 59th Annual Meeting of the Association
  for Computational Linguistics and the 11th International Joint Conference on
  Natural Language Processing (Volume 1: Long Papers)}, pages 908--921, Online.
  Association for Computational Linguistics.

\bibitem[{Keysers et~al.(2020)Keysers, Sch{\"a}rli, Scales, Buisman, Furrer,
  Kashubin, Momchev, Sinopalnikov, Stafiniak, Tihon, Tsarkov, Wang, van Zee,
  and Bousquet}]{keysers-etal-2020-measuring}
Daniel Keysers, Nathanael Sch{\"a}rli, Nathan Scales, Hylke Buisman, Daniel
  Furrer, Sergii Kashubin, Nikola Momchev, Danila Sinopalnikov, Lukasz
  Stafiniak, Tibor Tihon, Dmitry Tsarkov, Xiao Wang, Marc van Zee, and Olivier
  Bousquet. 2020.
\newblock \href {https://openreview.net/pdf?id=SygcCnNKwr} {Measuring
  compositional generalization: A comprehensive method on realistic data}.
\newblock In \emph{International Conference on Learning Representations
  (ICLR)}.

\bibitem[{Kim et~al.(2021)Kim, Ravikumar, Ainslie, and
  Ontañón}]{kim-etal-2021-improving}
Juyong Kim, Pradeep Ravikumar, Joshua Ainslie, and Santiago Ontañón. 2021.
\newblock \href {https://doi.org/10.18653/v1/2021.acl-short.81} {Improving
  compositional generalization in classification tasks via structure
  annotations}.
\newblock In \emph{Proceedings of the 59th Annual Meeting of the Association
  for Computational Linguistics and the 11th International Joint Conference on
  Natural Language Processing (Volume 2: Short Papers)}, pages 637--645,
  Online. Association for Computational Linguistics.

\bibitem[{Kim and Linzen(2020)}]{kim-linzen-2020-cogs}
Najoung Kim and Tal Linzen. 2020.
\newblock \href {https://doi.org/10.18653/v1/2020.emnlp-main.731} {{COGS}: A
  compositional generalization challenge based on semantic interpretation}.
\newblock In \emph{Proceedings of the 2020 Conference on Empirical Methods in
  Natural Language Processing}, pages 9087--9105, Online. Association for
  Computational Linguistics.

\bibitem[{Kingma and Ba(2015)}]{kingma-ba-2015-adam}
Diederik~P. Kingma and Jimmy Ba. 2015.
\newblock \href {http://arxiv.org/abs/1412.6980} {Adam: A method for stochastic
  optimization}.
\newblock \emph{Computing Research Repository (CoRR)}, arXiv:1412.6980.
\newblock Published as a conference paper at ICLR 2015.

\bibitem[{Kiperwasser and Goldberg(2016)}]{kiperwasser-goldberg-2016-simple}
Eliyahu Kiperwasser and Yoav Goldberg. 2016.
\newblock \href {https://doi.org/10.1162/tacl_a_00101} {Simple and accurate
  dependency parsing using bidirectional {LSTM} feature representations}.
\newblock \emph{Transactions of the Association for Computational Linguistics},
  4:313--327.

\bibitem[{Kitaev and Klein(2018)}]{kitaev-klein-2018-constituency}
Nikita Kitaev and Dan Klein. 2018.
\newblock \href {https://doi.org/10.18653/v1/P18-1249} {Constituency parsing
  with a self-attentive encoder}.
\newblock In \emph{Proceedings of the 56th Annual Meeting of the Association
  for Computational Linguistics (Volume 1: Long Papers)}, pages 2676--2686,
  Melbourne, Australia. Association for Computational Linguistics.

\bibitem[{Kuncoro et~al.(2018)Kuncoro, Dyer, Hale, Yogatama, Clark, and
  Blunsom}]{kuncoro-etal-2018-lstms}
Adhiguna Kuncoro, Chris Dyer, John Hale, Dani Yogatama, Stephen Clark, and Phil
  Blunsom. 2018.
\newblock \href {https://doi.org/10.18653/v1/P18-1132} {{LSTM}s can learn
  syntax-sensitive dependencies well, but modeling structure makes them
  better}.
\newblock In \emph{Proceedings of the 56th Annual Meeting of the Association
  for Computational Linguistics (Volume 1: Long Papers)}, pages 1426--1436,
  Melbourne, Australia. Association for Computational Linguistics.

\bibitem[{Lake and Baroni(2018)}]{lake-baroni-2018-generalization}
Brenden Lake and Marco Baroni. 2018.
\newblock \href {http://proceedings.mlr.press/v80/lake18a.html} {Generalization
  without systematicity: On the compositional skills of sequence-to-sequence
  recurrent networks}.
\newblock In \emph{Proceedings of the 35th International Conference on Machine
  Learning}, volume~80 of \emph{Proceedings of Machine Learning Research},
  pages 2873--2882, Stockholmsmässan, Stockholm Sweden. PMLR.

\bibitem[{Lewis et~al.(2020)Lewis, Liu, Goyal, Ghazvininejad, Mohamed, Levy,
  Stoyanov, and Zettlemoyer}]{lewis-etal-2020-bart-acl}
Mike Lewis, Yinhan Liu, Naman Goyal, Marjan Ghazvininejad, Abdelrahman Mohamed,
  Omer Levy, Veselin Stoyanov, and Luke Zettlemoyer. 2020.
\newblock \href {https://doi.org/10.18653/v1/2020.acl-main.703} {{BART}:
  Denoising sequence-to-sequence pre-training for natural language generation,
  translation, and comprehension}.
\newblock In \emph{Proceedings of the 58th Annual Meeting of the Association
  for Computational Linguistics}, pages 7871--7880, Online. Association for
  Computational Linguistics.

\bibitem[{Li et~al.(2017)Li, Xiong, Tu, Zhu, Zhang, and
  Zhou}]{li-etal-2017-modeling}
Junhui Li, Deyi Xiong, Zhaopeng Tu, Muhua Zhu, Min Zhang, and Guodong Zhou.
  2017.
\newblock \href {https://doi.org/10.18653/v1/P17-1064} {Modeling source syntax
  for neural machine translation}.
\newblock In \emph{Proceedings of the 55th Annual Meeting of the Association
  for Computational Linguistics (Volume 1: Long Papers)}, pages 688--697,
  Vancouver, Canada. Association for Computational Linguistics.

\bibitem[{Lindemann et~al.(2019)Lindemann, Groschwitz, and
  Koller}]{lindemann-etal-2019-compositional}
Matthias Lindemann, Jonas Groschwitz, and Alexander Koller. 2019.
\newblock \href {https://doi.org/10.18653/v1/P19-1450} {Compositional semantic
  parsing across graphbanks}.
\newblock In \emph{Proceedings of the 57th Annual Meeting of the Association
  for Computational Linguistics}, pages 4576--4585, Florence, Italy.
  Association for Computational Linguistics.

\bibitem[{Lindemann et~al.(2020)Lindemann, Groschwitz, and
  Koller}]{lindemann-etal-2020-fast}
Matthias Lindemann, Jonas Groschwitz, and Alexander Koller. 2020.
\newblock \href {https://doi.org/10.18653/v1/2020.emnlp-main.323} {Fast
  semantic parsing with well-typedness guarantees}.
\newblock In \emph{Proceedings of the 2020 Conference on Empirical Methods in
  Natural Language Processing (EMNLP)}, pages 3929--3951, Online. Association
  for Computational Linguistics.

\bibitem[{Linzen et~al.(2016)Linzen, Dupoux, and
  Goldberg}]{linzen2016assessing}
Tal Linzen, Emmanuel Dupoux, and Yoav Goldberg. 2016.
\newblock \href {https://doi.org/10.1162/tacl_a_00115} {Assessing the ability
  of {LSTM}s to learn syntax-sensitive dependencies}.
\newblock \emph{Transactions of the Association for Computational Linguistics},
  4:521--535.

\bibitem[{Liu et~al.(2021)Liu, An, Lin, Liu, Chen, Lou, Wen, Zheng, and
  Zhang}]{liu-etal-2021-learning}
Chenyao Liu, Shengnan An, Zeqi Lin, Qian Liu, Bei Chen, Jian-Guang Lou, Lijie
  Wen, Nanning Zheng, and Dongmei Zhang. 2021.
\newblock \href {https://aclanthology.org/2021.findings-acl.97} {Learning
  algebraic recombination for compositional generalization}.
\newblock In \emph{Findings of the Association for Computational Linguistics:
  ACL-IJCNLP 2021}, pages 1129--1144, Online. Association for Computational
  Linguistics.

\bibitem[{McCoy et~al.(2020)McCoy, Frank, and
  Linzen}]{DBLP:journals/tacl/McCoyFL20}
R.~Thomas McCoy, Robert Frank, and Tal Linzen. 2020.
\newblock \href {https://transacl.org/ojs/index.php/tacl/article/view/1892}
  {Does syntax need to grow on trees? sources of hierarchical inductive bias in
  sequence-to-sequence networks}.
\newblock \emph{Trans. Assoc. Comput. Linguistics}, 8:125--140.

\bibitem[{Orhan(2021)}]{orhan-2021-compgen}
A.~Emin Orhan. 2021.
\newblock \href {http://arxiv.org/abs/2109.15101} {Compositional generalization
  in semantic parsing with pretrained transformers}.
\newblock \emph{Computing Research Repository (CoRR)}, arXiv: 2109.15101.

\bibitem[{Partee(1984)}]{partee-1984-compositionality}
Barbara~H. Partee. 1984.
\newblock Compositionality.
\newblock In \emph{Varieties of Formal Semantics: Proceedings of the 4th
  Amsterdam Colloquium, September 1982}, volume~3, pages 281--311. Foris
  Publications, Dordrecht.

\bibitem[{Sennrich et~al.(2016)Sennrich, Haddow, and
  Birch}]{sennrich-etal-2016-controlling}
Rico Sennrich, Barry Haddow, and Alexandra Birch. 2016.
\newblock \href {https://doi.org/10.18653/v1/N16-1005} {Controlling politeness
  in neural machine translation via side constraints}.
\newblock In \emph{Proceedings of the 2016 Conference of the North {A}merican
  Chapter of the Association for Computational Linguistics: Human Language
  Technologies}, pages 35--40, San Diego, California. Association for
  Computational Linguistics.

\bibitem[{Shaw et~al.(2021)Shaw, Chang, Pasupat, and
  Toutanova}]{shaw-etal-2021-compositional}
Peter Shaw, Ming-Wei Chang, Panupong Pasupat, and Kristina Toutanova. 2021.
\newblock \href {https://aclanthology.org/2021.acl-long.75} {Compositional
  generalization and natural language variation: Can a semantic parsing
  approach handle both?}
\newblock In \emph{Proceedings of the 59th Annual Meeting of the Association
  for Computational Linguistics and the 11th International Joint Conference on
  Natural Language Processing (Volume 1: Long Papers)}, pages 922--938, Online.
  Association for Computational Linguistics.

\bibitem[{Tay et~al.(2021)Tay, Dehghani, Gupta, Aribandi, Bahri, Qin, and
  Metzler}]{tay-etal-2021-pretrained}
Yi~Tay, Mostafa Dehghani, Jai~Prakash Gupta, Vamsi Aribandi, Dara Bahri, Zhen
  Qin, and Donald Metzler. 2021.
\newblock \href {https://aclanthology.org/2021.acl-long.335} {Are pretrained
  convolutions better than pretrained transformers?}
\newblock In \emph{Proceedings of the 59th Annual Meeting of the Association
  for Computational Linguistics and the 11th International Joint Conference on
  Natural Language Processing (Volume 1: Long Papers)}, pages 4349--4359,
  Online. Association for Computational Linguistics.

\bibitem[{Tenney et~al.(2019)Tenney, Das, and Pavlick}]{tenney-etal-2019-bert}
Ian Tenney, Dipanjan Das, and Ellie Pavlick. 2019.
\newblock \href {https://doi.org/10.18653/v1/P19-1452} {{BERT} rediscovers the
  classical {NLP} pipeline}.
\newblock In \emph{Proceedings of the 57th Annual Meeting of the Association
  for Computational Linguistics}, pages 4593--4601, Florence, Italy.
  Association for Computational Linguistics.

\bibitem[{Vaswani et~al.(2017)Vaswani, Shazeer, Parmar, Uszkoreit, Jones,
  Gomez, Kaiser, and Polosukhin}]{vaswani-etal-2017-attention}
Ashish Vaswani, Noam Shazeer, Niki Parmar, Jakob Uszkoreit, Llion Jones,
  Aidan~N. Gomez, {\L}ukasz Kaiser, and Illia Polosukhin. 2017.
\newblock \href
  {https://proceedings.neurips.cc/paper/2017/file/3f5ee243547dee91fbd053c1c4a845aa-Paper.pdf}
  {Attention is all you need}.
\newblock In \emph{Advances in Neural Information Processing Systems 30 (NIPS
  2017)}. Curran Associates, Inc.

\bibitem[{Zheng and Lapata(2021)}]{zheng-lapata-2021-disentangled}
Hao Zheng and Mirella Lapata. 2021.
\newblock \href {http://arxiv.org/abs/2110.04655} {Disentangled sequence to
  sequence learning for compositional generalization}.
\newblock \emph{Computing Research Repository (CoRR)}, arXiv: 2110.04655.

\end{thebibliography}
\bibliographystyle{acl_natbib}

\newpage

\appendix

\section{COGS dataset statistics}\label{sec:cogsstatistics}
The COGS dataset 
contains English declarative sentences mapped with logical forms. It was created by \citet{kim-linzen-2020-cogs} and is publicly available at \url{https://github.com/najoungkim/COGS} (MIT license).
We use the version from April 2nd 2021 \href{https://github.com/najoungkim/COGS/tree/6f663835897945e94fd330c8cbebbdc494fbb690}{commit \texttt{6f66383}} and use the dataset as-is (no datapoints excluded or changed, use their data set splits), except for the AM parser for which we conduct the logical form to graph preprocessing described in \cref{subsec:cogs2graph}.\\
The normal training set (\sortof{train}) consists of 24,155 samples (24k in distribution, 143 primitives, 12 exposure examples), the dev and test set both contain 3k in distribution samples each.
Primitives and exposure examples contain \sortof{lexical trigger words} necessary for all but the three structural generalization types: these lexical trigger words each appear only once and in one sample in the whole training set.
Primitives are one-word sentences, therefore presenting word-meaning mapping without context of a sentence (necessary for the types Primitive to *). In contrast, exposure examples are full sentences e.g.~for the subject to object (common noun) generalization this sentence contains \word{hedgehog} as the subject. In the generalization set this word appears in 1k samples, but in a different syntactic configuration compared to the exposure example (e.g.~\word{hedgehog} in object position).
There is also an additional larger training set (\sortof{train100}) with 39,500 samples containing the lexical trigger words in 100 samples each, instead of just in one sample.
The out-of-distribution generalization set contains 21k samples, 1k per generalization type.

\section{Training details of the AM parser}
\label{sec:training_details_amparser}

\textit{The corresponding code will be made publicly available upon acceptance.}

\paragraph{Hyperparameters.}
For the AM parser, we mostly copied the hyperparameter values from the AMR experiments of \citet{groschwitz-etal-2021-learning}. This should help against overfitting on COGS, but we also note that hyperparameter tuning for compositional generalization datasets can be difficult anyways since one can typically easily achieve perfect scores on an in-doman dev set.
Copied values include for instance the number of epochs (60 due to supervised loss for edge existence and lexical labels), the batch size, the number and dimensionality of neural network layers and not using early stopping (but selecting best model based on per epoch evaluation metric on the dev set).
Choosing 3 sources has worked well on other datasets \citep{groschwitz-etal-2021-learning} and we adopt this hyperparameter choice. We note that with ditransitive verbs (i.e. verbs requiring NPs filling agent, theme, and recipient roles) present in COGS we need at least three sources anyway to account for these.

\paragraph{Deviations from \citet{groschwitz-etal-2021-learning}'s settings.}
For training on train (but not train100), we set the vocabulary threshold from 7 down to 1 to account for the fact that the lexical generalizations rely on a single occurrence of a word in the training data (on train100 we keep 7 as a threshold since the trigger words occur 100 times in there). 
Furthermore, the COGS dataset doesn't have part-of-speech, lemma or named-entity annotations, so we just don't use embeddings for these.
For the word embeddings we either use BERT-Large-uncased \citep{devlin-etal-2019-bert} or learn embeddings from the dataset only (embedding dimension 1024, same as for the BERT model).
We also decreased the learning rate from 0.001 to 0.0001: we observed that the learning curves are still converging very quickly and hypothesize that COGS training set might also be easier than the AMR one used in \citet{groschwitz-etal-2021-learning}.\\
Unlike them we didn't use the fixed-tree decoder (described in \citealt{groschwitz-etal-2018-amr}), but opted for the projective A* decoder \citep[§4.2]{lindemann-etal-2020-fast}: in pre-experiments this showed better results. In addition, it makes comparison to related work (such as LeAR by \citet{liu-etal-2021-learning}) easier which uses only projective latent trees. 
We also use supervised loss for edge existence and lexical labels: we can use supervised loss for both as they do not depend on the source names to be learnt. In preliminary experiments this yielded better results than using the automaton-based loss for them too. The supervised loss wasn't described in \citet{groschwitz-etal-2021-learning}, but already implemented in their code base and they note there that the effect on performance was mixed in their experiments (similar for SDP, worse for AMR).

\paragraph{Relative distance encoding.} 
For the relative distance encodings we added to the dependency edge existence scoring, we used sine-cosine interleaved encoding function introduced by \citet[§3.5]{vaswani-etal-2017-attention} and as input to it use the relative distance $dist(i,j)=i-j$ between sentence positions $i$ and $j$. We use a dimensionality of 64 for the distance encodings ($d_{model}$ in \citet{vaswani-etal-2017-attention} is 512). These distance encodings are then concatenated together with the BiLSTM representations for possible heads and dependents used in the standard \citet{kiperwasser-goldberg-2016-simple} edge scoring model. This constitutes the input to the MLP emitting a score for each token pair. In other words, for each token pair $\langle i,j\rangle$ the MLP has to decide edge existence based on the representations of the tokens at positions $i$ and $j$, and an encoding of the relative distance $dist(i,j)=i-j$.
These models have the suffix \sortof{\dist} in the tables.


\paragraph{Runtimes.}
Training the AM parser took 5 to 7 hours on train with 60 epochs and 6 to 9.5 hours on train100. In general, training with BERT took longer than without, same holds for adding relative distance encodings.
Inference with a trained model on the full 21k generalization samples took about 15 minutes using the Astar decoder with the \sortof{ignore aware} heuristic.
All AM parser experiments were performed using Intel Xeon E5-2687W v3 10-core processors at 3.10Ghz and 256GB RAM, and MSI Nvidia Titan-X (2015) GPU cards (12GB). 

\paragraph{Number of parameters.}
For their models, \citet{kim-linzen-2020-cogs} tried to keep the number of parameters comparable (9.5 to 11 million) and therefore rule out model capacity as a confound. The number of trainable parameters of the AM parser model used is 10.7 to 11.5million (lower one is with BERT, higher without. Impact of relative distance encoding is rather minimal: $<17$k), so the improved performance is not just due to a higher number of parameters.


\paragraph{Dev set performance.}
As usual for compositional generalization datasets, it is relatively easy to get (near) perfect results on the (in domain) dev/test sets. We observed this too: all AM parser models had an exact match score of at least 99.9 on the dev set and at least 99.8 on the (in distribution) test set. 

\paragraph{Evaluation procedure.}
Unfortunately, \citet{kim-linzen-2020-cogs} didn't provide a separate evaluation script. 
As a main evaluation metric they use (string) exact match accuracy on the logical forms which we adopt. 
Note that this requires models to learn the \sortof{correct} order of conjuncts: even if a logically equivalent form with a different order of conjuncts would be predicted, string exact match would count it as a failure.
In lack of an official evaluation script we implemented our own evaluation script to compute exact match. 


\section{Training details of Seq2seq} \label{sec:training_details_bart}

\paragraph{Hyperparameters.}  We use the same hyperparameter setting for BART on both syntactic and semantic experiments. 
We use \textit{bart-base}\footnote{\url{https://huggingface.co/facebook/bart-base}} model in all our experiments. Our batch size is 64.
We use Adam optimizer \cite{kingma-ba-2015-adam} with learning rate 1e-4 and
gradient accumulation steps 8. Loss averaged over tokens is used as the validation metric
for early stopping following \citet{kim-linzen-2020-cogs}.  During
inference, we use beam search with beam size 4.

\paragraph{Dev set performance.} The exact match accuracy is at least 99.6 for both dev set and (in-distribution) test set in all experiments. 

\paragraph{Other details.} Training took 4 hours for BART with about 80 epochs on train and 5 hours with about 50 epochs on train100. Inference on generalization set took about 1 hour. All BART experiments were run on Tesla V100 GPU cards (32GB). The number of parameters in our BART model is 140  million. 

\paragraph{Syntactic annotations.} To obtain syntactic annotations, we use NLTK\footnote{\url{https://www.nltk.org/}} to parse each sentence in COGS with PCFG grammar generating COGS. In our experiments, we found this parsing process did not yield any ambiguous tree. The original PCFG grammar contains rules such as \lform{NP$\to$NP\_animate\_dobj\_noPP}. We replace such fine-grained nonterminals (e.g.\ \lform{NP\_animate\_dobj\_noPP}) with general nonterminals (e.g.\ \lform{NP}). This results in duplicate patterns (e.g.\ \lform{NP$\to$NP}) and we further remove such patterns from the output tree. 

\paragraph{Results from other papers.} \citet{conklin-etal-2021-meta}\footnote{\url{https://github.com/berlino/tensor2struct-public}}, \citet{akyurek-andreas-2021-lexicon}\footnote{\url{https://github.com/ekinakyurek/lexical}}, \citet{csordas-etal-2021-devil}\footnote{\url{https://github.com/robertcsordas/transformer_generalization}} and \citet{tay-etal-2021-pretrained} did not report performance of their model on train100 set. To report these numbers, we additionally use their published code to train their model on train100 for 5 runs. We use seed 6-10 for \citet{conklin-etal-2021-meta} and random number seeds for \citet{csordas-etal-2021-devil}, following their default setting. We use their default configuration file for their best model to set the hyperparameters. \citet{tay-etal-2021-pretrained}, did not publish their code so we did not report that. \citet{orhan-2021-compgen}\footnote{\url{https://github.com/eminorhan/parsing-transformers}} and \citet{zheng-lapata-2021-disentangled} are the two most recently published seq2seq approaches. Both did not provide numbers for train100 training and because of their recency we weren't able to run their models on the train100 set so far. We thus only report their published results for train set. 




\section{Detailed evaluation results}\label{sec:detailed_evaluation_results}


The main results are summarized in the main paper in \cref{subsec:results} with \cref{tab:results} and \cref{tab:selected_gentype_eval}.
Here we present 
AM parser (\cref{tab:detailed_eval_accuracy}), 
LeAR (\cref{tab:detailed_eval_accuracy_lear}) and BART (\cref{tab:detailed_eval_accuracy_seq2seq})
 performance for each of COGS' 21 generalization types separately with the usual mean and standard deviation of 5 runs.
For descriptions of the generalization types we refer to \citet[][§3 and Fig.~1]{kim-linzen-2020-cogs}.

\paragraph{On accuracy computation for LeAR.}
We observed that the LeAR model skips 22 sentences in the generalization set due to out-of-vocabulary tokens.\footnote{The words \word{gardner} and \word{monastery} occur zero times in the train set, but in total in 22 sentences of the generalization set. The majority (15) of these appear in PP recursion samples.}
We do include these sentences in the accuracy computation (as failures) for the generalization set. 
The published LeAR code does not convert its internally used representation back to logical forms, therefore we evaluate on the logical forms like it is done for other models, but have to rely on accuracy computation done in the LeAR code for the internal representation.
Furthermore we would like to note that--based on inspecting the published code\footnote{\url{https://github.com/thousfeet/LEAR} }--, LeAR made the preprocessing choice to ignore the contribution of the definite determiner, basically treating indefinite and definite NPs equally, resulting in a big conjunction without any iota (\sortof{\lform{*}}) prefixes. 

\paragraph{On model numbers copied from other papers.}
\citet{kim-linzen-2020-cogs} provide three baseline models, among which the Transformer model reached the best performance on train and train100. Per generalization type results can be found in their Appendix F (Table 5 on page 9105) from which we report the Transformer model numbers.\\
The strongest model of \citet{akyurek-andreas-2021-lexicon} is actually \sortof{Lex:Simple:Soft} (cf. their Table~5) with a generalization accuracy of 83\% (also reported in our \cref{tab:results}), whereas their Lex:Simple model lags 1 point behind. For the latter, but not for the former, the authors provide per generalization type output in their accompanying GitHub repository as part of a \href{https://github.com/ekinakyurek/lexical/blob/e7a44e19d23a1d99726cd76c5cd88f56ca586653/analyze.ipynb}{jupyter notebook}. Therefore numbers in \cref{tab:selected_gentype_eval} are for Lex:Simple, not Lex:Simple:Soft.\\
We picked the best performing model of \citet{orhan-2021-compgen}: According to their Table 2 the \verb+t5-3b mt5_xl+ model shows the best generalization performance (84.6\% average accuracy). From the accompanying GitHub repository\footnote{\url{https://github.com/eminorhan/parsing-transformers}} we copy the model's results, specifically we average over the 5 runs of the model \href{https://github.com/eminorhan/parsing-transformers/tree/9887632a348f9d2e3b010f86a7931691a0faf044/results/3b/cogs_mt5/epochs_10}{3b-cogs-mt5-epochs10 (commit \texttt{04a2508})}. We note that other models reported in \citet{orhan-2021-compgen} showed the same performance pattern with respect to our three generalization classes \lexg, \propg, and \structg.\\
%
For \citet{zheng-lapata-2021-disentangled}, our reported number is slightly different from the original paper. This is because we asked the authors for detailed results and they provide us with their newest results averaged over 5 runs.

\paragraph{Abbreviations in the tables.} 
\sortof{Subj} means \sortof{subject}, 
\sortof{Obj} means \sortof{object}, 
\sortof{Prim} means \sortof{primitive}, 
\sortof{Infin. arg} means \sortof{infinitival argument},
\sortof{ObjmodPP to SubjmodPP} means \sortof{object-modifying PP to subject-modifying PP},
\sortof{ObjOTrans.} means \sortof{object omitted transitive},
\sortof{trans.} means \sortof{transitive},
\sortof{unacc} means \sortof{unaccusative},
\sortof{Dobj} means \sortof{Double Object}.

\begin{table*}[tbh]
    \tiny
    \centering
\begin{tabular}{l|S[table-format=2.1,table-auto-round=true]@{$\pm$}S@{\hspace{4pt}}S[table-format=2.1,table-auto-round=true]@{$\pm$}S@{\hspace{4pt}}S[table-format=2.1,table-auto-round=true]@{$\pm$}S@{\hspace{4pt}}S[table-format=2.1,table-auto-round=true]@{$\pm$}S|S[table-format=2.1,table-auto-round=true]@{$\pm$}S@{\hspace{4pt}}S[table-format=2.1,table-auto-round=true]@{$\pm$}S@{\hspace{4pt}}S[table-format=2.1,table-auto-round=true]@{$\pm$}S@{\hspace{4pt}}S[table-format=2.1,table-auto-round=true]@{$\pm$}S}
 & \multicolumn{8}{c|}{{train}} & \multicolumn{8}{c}{{train100}} \\ \cmidrule(lr){2-9} \cmidrule(lr){10-17}
Type & \multicolumn{2}{l}{\amToken} & \multicolumn{2}{l}{\amToken+\dist} & \multicolumn{2}{l}{\amBert} & \multicolumn{2}{l}{\amBert+\dist} & \multicolumn{2}{l}{\amToken} & \multicolumn{2}{l}{\amToken+\dist} & \multicolumn{2}{l}{\amBert} & \multicolumn{2}{l}{\amBert+\dist} \\ \midrule
Subj to Obj (common noun) & 65.78 & 43.4 & 88.34 & 10.9 & 99.68 & 0.1 & 96.52 & 6.8 & 99.90 & 0.1 & 99.90 & 0.1 & 99.96 & 0.1 & 99.92 & 0.2 \\
Subj to Obj (proper noun) & 69.94 & 9.8 & 48.06 & 32.0 & 66.32 & 38.8 & 61.80 & 47.3 & 98.90 & 1.7 & 100.00 & 0.0 & 89.56 & 8.1 & 95.82 & 9.3 \\
Obj to Subj (common noun) & 53.08 & 45.0 & 97.94 & 4.4 & 99.90 & 0.2 & 88.00 & 26.7 & 99.90 & 0.1 & 99.82 & 0.2 & 99.96 & 0.1 & 99.94 & 0.1 \\
Obj to Subj (proper noun) & 90.04 & 21.4 & 88.26 & 25.9 & 88.88 & 11.2 & 78.84 & 42.9 & 99.82 & 0.0 & 99.84 & 0.1 & 99.88 & 0.0 & 99.90 & 0.0 \\ 
\midrule
Prim to Subj (common noun) & 3.42 & 7.6 & 0.00 & 0.0 & 76.22 & 42.2 & 80.30 & 42.2 & 97.96 & 4.5 & 59.90 & 54.7 & 100.00 & 0.0 & 100.00 & 0.0 \\
Prim to Subj (proper noun) & 4.72 & 10.6 & 1.04 & 2.3 & 99.94 & 0.1 & 100.00 & 0.0 & 99.76 & 0.3 & 99.90 & 0.1 & 99.98 & 0.0 & 99.96 & 0.1 \\
Prim to Obj (common noun) & 0.18 & 0.4 & 0.02 & 0.0 & 74.46 & 32.5 & 80.06 & 40.7 & 95.88 & 8.9 & 59.92 & 54.7 & 100.00 & 0.0 & 100.00 & 0.0 \\
Prim to Obj (proper noun) & 10.38 & 9.1 & 22.04 & 15.6 & 90.48 & 9.9 & 94.92 & 3.7 & 98.82 & 2.4 & 99.82 & 0.4 & 84.90 & 9.1 & 94.44 & 9.0 \\
Prim verb to Infin. arg & 59.72 & 54.2 & 55.16 & 50.5 & 100.00 & 0.0 & 82.90 & 38.2 & 17.56 & 30.8 & 1.00 & 2.2 & 100.00 & 0.0 & 100.00 & 0.0 \\
\midrule
ObjmodPP to SubjmodPP & 38.14 & 23.1 & 26.08 & 15.1 & 59.04 & 40.8 & 71.50 & 24.0 & 48.00 & 17.3 & 44.78 & 23.9 & 49.08 & 27.5 & 77.70 & 7.1 \\
CP recursion & 100.00 & 0.0 & 99.96 & 0.1 & 100.00 & 0.0 & 100.00 & 0.0 & 99.94 & 0.1 & 99.98 & 0.0 & 100.00 & 0.0 & 99.98 & 0.0 \\
PP recursion & 60.52 & 4.2 & 97.58 & 0.9 & 36.34 & 8.0 & 97.32 & 2.0 & 57.18 & 8.3 & 96.96 & 1.1 & 41.48 & 11.2 & 98.62 & 0.5 \\
\midrule
Active to Passive & 69.32 & 42.2 & 41.66 & 52.3 & 83.02 & 24.8 & 78.82 & 31.3 & 100.00 & 0.0 & 100.00 & 0.0 & 100.00 & 0.0 & 100.00 & 0.0 \\
Passive to Active & 51.56 & 45.2 & 46.62 & 50.2 & 45.54 & 27.2 & 51.96 & 43.6 & 99.62 & 0.7 & 99.94 & 0.1 & 99.98 & 0.0 & 100.00 & 0.0 \\
ObjOTrans. to trans. & 79.56 & 33.6 & 77.82 & 28.2 & 22.26 & 24.0 & 35.56 & 33.4 & 99.94 & 0.1 & 99.96 & 0.1 & 99.98 & 0.0 & 100.00 & 0.0 \\
Unacc to transitive & 33.16 & 36.1 & 51.24 & 47.2 & 48.16 & 35.8 & 48.86 & 41.5 & 99.60 & 0.7 & 99.96 & 0.1 & 99.98 & 0.0 & 100.00 & 0.0 \\
Dobj dative to PP dative & 99.32 & 0.8 & 98.76 & 2.0 & 99.80 & 0.1 & 95.00 & 11.0 & 99.94 & 0.1 & 99.90 & 0.1 & 99.98 & 0.0 & 99.98 & 0.0 \\
PP dative to Dobj dative & 90.38 & 11.9 & 79.52 & 44.5 & 85.60 & 21.7 & 89.46 & 11.5 & 99.68 & 0.1 & 99.80 & 0.1 & 100.00 & 0.0 & 99.98 & 0.0 \\ 
\midrule
Agent NP to Unacc Subj & 78.52 & 43.4 & 99.66 & 0.6 & 95.32 & 6.4 & 78.20 & 43.9 & 100.00 & 0.0 & 100.00 & 0.0 & 100.00 & 0.0 & 100.00 & 0.0 \\
Theme NP to ObjOTrans. Subj & 99.94 & 0.1 & 99.24 & 1.7 & 99.92 & 0.1 & 70.48 & 41.9 & 100.00 & 0.0 & 100.00 & 0.0 & 100.00 & 0.0 & 100.00 & 0.0 \\
Theme NP to Unergative Subj & 99.96 & 0.1 & 96.60 & 7.6 & 99.94 & 0.1 & 64.36 & 49.0 & 100.00 & 0.0 & 100.00 & 0.0 & 100.00 & 0.0 & 100.00 & 0.0 \\ 
\midrule
Total & 59.89 & 21.1 & 62.65 & 18.7 & 79.56 & 15.4 & 78.33 & 27.7 & 91.07 & 3.6 & 88.59 & 6.6 & 93.56 & 2.7 & 98.39 & 1.3 \\
\bottomrule
    \end{tabular}
    \caption{
        Exact match accuracy on the generalization set by generalization type for all AM parser models. 
    }\label{tab:detailed_eval_accuracy}
\end{table*}

\begin{table}[tbh]
    \tiny
    \centering
\begin{tabular}{lS[table-format=2.1,table-auto-round=true]@{$\pm$}S}
 & \multicolumn{2}{c}{{train}} \\
Type & \multicolumn{2}{c}{LeAR} \\ \midrule
Subj to Obj (common noun) & 99.8 & 0.0 \\
Subj to Obj (proper noun) & 93.1 & 10.2 \\
Obj to Subj (common noun) & 100.0 & 0.0 \\
Obj to Subj (proper noun) & 99.9 & 0.0  \\ 
\midrule
Prim to Subj (common noun) & 100.0 & 0.0 \\
Prim to Subj (proper noun) & 100.0 & 0.0 \\
Prim to Obj (common noun) & 99.8 & 0.0 \\
Prim to Obj (proper noun) & 93.1 & 10.2 \\
Prim verb to Infin. arg & 100.0 & 0.0 \\
\midrule
ObjmodPP to SubjmodPP & 92.5 & 9.4 \\
CP recursion & 100.00 & 0.0  \\
PP recursion & 98.5 & 0.0 \\
\midrule
Active to Passive & 100.0 & 0.0 \\
Passive to Active & 100.0 & 0.0 \\
ObjOTrans. to trans. & 100.0 & 0.0 \\
Unacc to transitive & 100.0 & 0.0 \\
Dobj dative to PP dative & 99.9 & 0.0 \\
PP dative to Dobj dative & 90.9 & 0.0 \\ 
\midrule
Agent NP to Unacc Subj & 100.0 & 0.0 \\
Theme NP to ObjOTrans. Subj & 100.0 & 0.0 \\
Theme NP to Unergative Subj & 100.0 & 0.0 \\ 
\midrule
Total & 98.8809 & 0.9 \\
\bottomrule
    \end{tabular}
    \caption{
        Exact match accuracy on the generalization set by generalization type for the LeAR reproduction runs on train.
    }\label{tab:detailed_eval_accuracy_lear}
\end{table}

\begin{table*}[tbh]
    \tiny
    \centering
\begin{tabular}{l|S[table-format=2.1,table-auto-round=true]@{$\pm$}S@{\hspace{6pt}}S[table-format=2.1,table-auto-round=true]@{$\pm$}S@{\hspace{4pt}}|S[table-format=2.1,table-auto-round=true]@{$\pm$}S@{\hspace{4pt}}S[table-format=2.1,table-auto-round=true]@{$\pm$}S@{\hspace{4pt}}S[table-format=2.1,table-auto-round=true]@{$\pm$}S@{\hspace{4pt}}S[table-format=2.1,table-auto-round=true]@{$\pm$}S@{\hspace{4pt}}}
 & \multicolumn{4}{c}{{train}} & \multicolumn{8}{c}{{train100}} \\ \cmidrule(lr){2-5} \cmidrule(lr){6-13}
Type & \multicolumn{2}{l}{BART} & \multicolumn{2}{l}{BART+syn} & \multicolumn{2}{l}{BART} & \multicolumn{2}{l}{BART+syn} & \multicolumn{2}{l}{BART+mtl} & \multicolumn{2}{l}{BART+mask} \\ \midrule
Subj to Obj (common noun) & 98.6 & 0.8 & 99.6 & 0.2 & 99.2 & 0.2 & 99.8 & 0.1 & 99.6 & 0.1 & 92.9 & 1.2 \\
Subj to Obj (proper noun) & 68.7& 1.1 & 87.0 & 3.2 & 85.7 & 6.7 & 94.7 & 5.3 &  80.1 & 5.5 & 93.3 & 3.4 \\
Obj to Subj (common noun) & 99.2 & 0.6 & 99.8 & 0.1 & 99.1 & 1.3 & 99.7 & 0.1 & 99.6 & 0.2 & 98.7 & 0.4 \\
Obj to Subj (proper noun) & 99.4 & 0.4 & 99.8 & 0.0 & 99.5 & 0.2 & 99.8 & 0.1 & 97.8 & 1.5 & 99.3 & 0.3 \\
\midrule
Prim to Subj (common noun) & 98.4 & 1.3 & 99.9 & 0.0 & 95.0 & 9.0 & 99.9 & 0.0 & 99.7 & 0.0 & 99.6 & 0.2 \\
Prim to Subj (proper noun) & 98.6 & 0.9 & 100.0 & 0.1 & 95.5 & 4.3 & 100.0 & 0.0 &  99.9 & 0.1 & 98.9 & 1.1 \\
Prim to Obj (common noun) & 98.9 & 0.6 & 99.5 & 0.2 & 99.4 & 0.2 & 99.8 & 0.0 &  99.6 & 0.1 & 96.1 & 0.9 \\
Prim to Obj (proper noun) & 65.2 & 4.4 & 88.6 & 4.3 &55.2 & 27.1 & 98.1 & 2.1 &  94.6 & 0.3 & 94.8 & 2.0 \\
Prim verb to Infin. arg & 99.9 & 0.1 & 100.0 & 0.0 & 100.0 & 0.0 & 100.0 & 0.0 & 100.0 & 0.0 &  99.9 & 0.0 \\
\midrule
ObjmodPP to SubjmodPP & 0.0 & 0.0 & 0.0 & 0.0 & 0.0 & 0.0 & 0.0 & 0.0 & 0.0 & 0.0 &  0.0 & 0.0 \\
CP recursion & 0.3 & 0.3 & 5.9 & 1.2 & 0.2 & 0.4 &  6.5 & 0.5 & 0.2 & 0.2 & 1.1 & 0.5 \\
PP recursion & 11.2 & 1.7 & 6.7 & 0.2 & 10.2 & 1.8 &  7.5 & 0.4 & 11.7 & 0.3 & 10.6 & 1.4 \\
\midrule
Active to Passive & 99.9 & 0.0 & 99.9 & 0.0 & 99.9 & 0.0 & 99.9 & 0.0 & 100.0 & 0.0 &  99.9 & 0.0 \\
Passive to Active & 99.5 & 0.2 & 99.9 & 0.0 & 99.9 & 0.0 & 99.9 & 0.0 & 99.9 & 0.1 &  99.8 & 0.2 \\
ObjOTrans. to trans. & 99.6 & 0.3 & 100.0 & 0.0 & 99.9 & 0.1 & 100.0 & 0.0 & 99.9 & 0.1 & 99.9 & 0.2 \\
Unacc to transitive & 0.0 & 0.0 & 0.0 & 0.0 & 99.9 & 0.0 & 100.0 & 0.0 & 99.9 & 0.1 & 99.7 & 0.2 \\
Dobj dative to PP dative & 98.3 & 1.2 & 99.4 & 0.3 & 99.2 & 0.2 & 99.5 & 0.2 &  99.3 & 0.0 & 99.1 & 0.1 \\
PP dative to Dobj dative & 98.6 & 1.6 & 99.8 & 0.0 & 99.5 & 0.1 &  99.9 & 0.1 & 99.6 & 0.2 & 99.2 & 0.3 \\
\midrule
Agent NP to Unacc Subj & 96.2 & 1.4 & 99.1 & 1.0 & 99.8 & 0.2 & 99.6 & 0.3 & 100.0 & 0.0 & 96.2 & 0.9 \\
Theme NP to ObjOTrans. Subj & 98.8 & 0.8 & 99.8 & 0.3 & 99.6 & 0.2 & 99.9 & 0.0 & 100.0 & 0.0 & 92.5 & 5.5 \\
Theme NP to Unergative Subj & 99.1 & 0.7 & 99.8 & 0.3 & 99.8 & 0.2 &  99.8 & 0.1 &  100.0 & 0.0 &  94.1 & 4.1 \\
\midrule
Total & 77.5 & 0.4 & 80.2 & 0.4 & 82.7 & 1.3 & 85.9 & 0.3 & 84.8 & 0.2 & 84.1 & 0.4 \\

\bottomrule
    \end{tabular}
    \caption{
        Exact match accuracy on the generalization set by generalization type for all BART models. 
    }\label{tab:detailed_eval_accuracy_seq2seq}
\end{table*}

\section{Additional information on COGS to graph conversions}\label{sec:details_cogs2graph}


This is a more detailed explanation of the COGS logical form to graph conversion described in \cref{subsec:cogs2graph}
based on four additional example sentences:
\begin{exe}
    \ex\label{ex:boywanted} The boy wanted to go. \\
    \lform{*boy($x_1$); want.agent($x_2, x_1$) $\land$ want.xcomp($x_2, x_4$) \\ $\land$ go.agent($x_4, x_1$)}
    \ex\label{ex:other} Ava was lended a cookie in a bottle. \\
    \lform{lend.recipient($x_2,$ Ava) \\ $\land$ lend.theme($x_2, x_4$) \\ $\land$ cookie($x_4$) \\ $\land$ cookie.nmod.in($x_4, x_7$) \\ $\land$ bottle($x_7$)}
    \ex\label{ex:cprec} Ava said that Ben declared that Claire slept. \\
    \lform{say.agent($x_1,$ Ava) \\ $\land$ say.ccomp($x_1, x_4$) \\ $\land$ declare.agent($x_4,$ Ben) \\ $\land$ declare.ccomp($x_4, x_7$) \\ $\land$  sleep.agent($x_7,$ Claire)}
    \ex\label{ex:prim_touch} touch \\
    \lform{$\lambda a. \lambda b. \lambda e.$ touch.agent($e, b$) $\land$ touch.theme($e, a$)}
\end{exe}
The first of these is used as the main example for now. Its graph conversion can be found in \cref{fig:cogsgraph-boy}.
\begin{figure}[t] 
    \centering
\begin{tikzpicture}[node distance=30pt,font=\scriptsize]
    \node (want) [nnode, very thick] {$x_2$ / want};
    \node (the) [nnode, left of=want, xshift=-30pt] {$x_0$ / the};
    \node (boy) [nnode, below of=the] {$x_1$ / boy};
    \node (go) [nnode, right of=boy, xshift=+30pt] {$x_4$ / go};
    \draw [arrow] (want) -- node[anchor=south] {agent} (boy);
    \draw [arrow] (want) -- node[anchor=west] {xcomp} (go);
    \draw [arrow] (go) -- node[anchor=north] {agent} (boy);
    \draw [arrow] (the) -- node[anchor=east] {iota} (boy);
\end{tikzpicture}\\[.5pt]
{\footnotesize
\lform{* boy($x_1$) ; want.agent($x_2, x_1$) $\land$ want.xcomp($x_2, x_4$) $\land$ go.agent($x_4, x_1$)}
}
    \caption{
        Logical form to graph conversion for \sentence{The boy wanted to go} (cf.~\cref{ex:boywanted}).
        For illustration only we use node names (the part before the `/') to outline the token alignment.
    }\label{fig:cogsgraph-boy}
\end{figure}

\paragraph{Basic ideas.} 
\emph{Arguments} of predicates (variables like $x_i$ or proper names like \lform{Ava}) are translated to nodes. 
The first part of each predicate name (e.g.~\lform{boy}, \lform{want}, \lform{go}) is the lemma of the token pointed to by the first argument (e.g.~$x_1, x_2, x_4$), we strip this lemma (\sortof{delexialize}) from the predicate and insert it as the node label of the first argument (post-processing reverses this).\\
\emph{Binary predicates} (i.e.~terms with 2 arguments) are translated into edges, pointing from their first to their second argument, e.g.~\lform{want.agent($x_2, x_1$)} is converted to an \graphedge{agent} edge from node $x_2$ (the \graphnode{want} node) to node $x_1$. 
Because of the delexicalization described above, there are only 8 different edge labels: \graphedge{agent}, \graphedge{theme}, \graphedge{recipient}, \graphedge{xcomp}, \graphedge{ccomp}, \graphedge{iota} and 2 preposition-introduced edges described below.\\
For \emph{unary predicates} like \lform{boy($x_1$)} the delexicalization already suffices, so we don't add any edge (in lack of a proper target node). We restore unary predicates during postprocessing for nodes with no outgoing edges.\\
Each \emph{iota term} \lform{*noun($x_{\text{noun}}$);} is treated as if it was a conjunction of the noun meaning (i.e.~\lform{noun($x_{\text{noun}}$)}) and \sortof{definite determiner meaning} binary predicate \lform{the.iota($x_{\text{the}}, x_{\text{noun}}$)}.\\
The AM parser further requires one node to be the \emph{root node}. 
For non-primitives we select it heuristically as the node with no incoming edges (excluding preposition and determiner nodes).

\begin{figure}[tbh]
    \centering
    \begin{subfigure}[b]{\columnwidth} 

\begin{tikzpicture}[node distance=1.3cm,font=\small]
    \node (lend) [nnode, very thick] {$x_2$ / lend};
    \node (ava) [nnode, below of=lend, xshift=-25pt] {$x_0$ / Ava};
    \node (cookie) [nnode, right of=ava, xshift=+25pt] {$x_4$ / cookie};
    \node (in)     [nnode, right of=lend, xshift=+40pt] {$x_5$ / in};
    \node (bottle) [nnode, right of=cookie, xshift=+50pt] {$x_7$ / bottle};
    \draw [arrow] (lend) -- node[anchor=east] {recipient} (ava);
    \draw [arrow] (lend) -- node[anchor=east] {theme} (cookie);
    \draw [arrow] (in) -- node[anchor=west] {nmod.op1} (cookie.north);
    \draw [arrow] (in) -- node[anchor=west] {nmod.op2} (bottle.north);
\end{tikzpicture}
        \caption{See also \cref{ex:other}.}\label{fig:lendedCookie}
    \end{subfigure}
    \begin{subfigure}[b]{\columnwidth} 
\begin{tikzpicture}[node distance=1.3cm,font=\small]
    \node (say) [nnode, very thick] {$x_1$ / say};
    \node (ava) [nnode, below of=say] {$x_0$ / Ava};
    \node (declare) [nnode, right of=say, xshift=+40pt] {$x_4$ / declare};
    \node (ben) [nnode, below of=declare] {$x_3$ / Ben};
    \node (sleep) [nnode, right of=declare, xshift=+40pt] {$x_7$ / sleep};
    \node (claire) [nnode, below of=sleep] {$x_6$ / Claire};
    \draw [arrow] (say) -- node[anchor=west] {agent} (ava);
    \draw [arrow] (say) -- node[anchor=south] {ccomp} (declare);
    \draw [arrow] (declare) -- node[anchor=west] {agent} (ben);
    \draw [arrow] (declare) -- node[anchor=south] {ccomp} (sleep);
    \draw [arrow] (sleep) -- node[anchor=west] {agent} (claire);
\end{tikzpicture}
        \caption{See also \cref{ex:cprec}.}\label{fig:cprecconversion}
    \end{subfigure}
    \begin{subfigure}[b]{.5\columnwidth} 
\begin{tikzpicture}[node distance=1.3cm,font=\small]
    \node (touch) [nnode, very thick] {$e_0$ / touch};
    \node (b) [nnode, below of=touch, xshift=-25pt] {$b_0$ / \textcolor{red}{\src{s0}}};
    \node (a) [nnode, below of=touch, xshift=+25pt] {$a_0$ / \textcolor{red}{\src{s1}}};
    \draw [arrow] (touch) -- node[anchor=east] {agent} (b);
    \draw [arrow] (touch) -- node[anchor=west] {theme} (a);
\end{tikzpicture}
        \caption{See also \cref{ex:prim_touch}.}\label{fig:prim_touch}
    \end{subfigure}
    \caption{
        Results of the logical form to graph conversion for \crefrange{ex:other}{ex:prim_touch}.
        Actually for (\subref{fig:prim_touch}) the tree automaton contained all possible source name combinations for nodes $a$ and $b$, not just $\langle$\textcolor{red}{\src{s0}},\textcolor{red}{\src{s1}}$\rangle$.
    }\label{fig:moreexamples}
\end{figure}
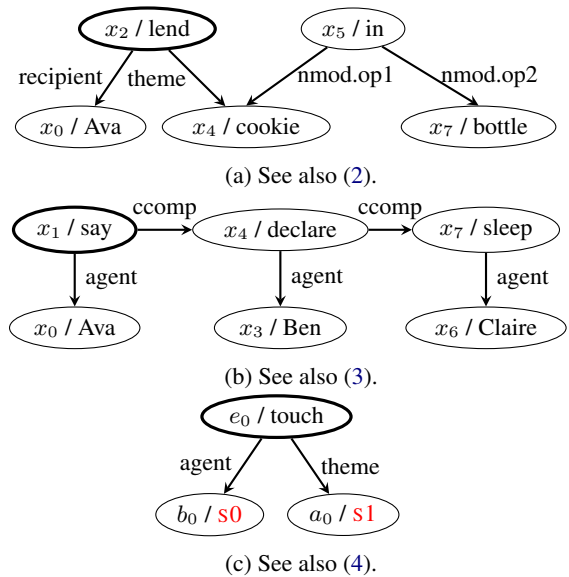

\paragraph{Prepositions.}
Instead of being treated as an edge as the above would suggest, 
we \sortof{reify} them, 
so each preposition becomes a node of the graph with outgoing \graphedge{nmod} edges to the modified NP and the argument NP. So for \word{cookie in the bottle} (cf. \cref{ex:other} and \cref{fig:lendedCookie}) we create a node with label \graphedge{in} and draw an outgoing \graphedge{nmod.op1} edge to the \graphnode{cookie}-node and an \graphedge{nmod.op2} edge to the \graphnode{bottle}-node.

\paragraph{Alignments.} 
For training the AM parser additionally needs \emph{alignments} of the nodes to the input tokens.
Luckily all $x_i$ nodes naturally provide alignments (alignment to $i$th input token). 
For proper names we simply align them to the first occurrence in the sentence\footnote{
    this works because it seems that a name never appears more than once within a sentence.
    Names in the logical forms also seem to be ordered based on their token position.
}, the special determiner node is aligned to the token preceding the corresponding $x_{noun}$.\footnote{
    we can do so because there are --beyond \word{the} and \word{a}-- no pre-nominal modifiers like adjectives in this dataset.
}
The edges are implicitly aligned by the blob heuristics, which are pretty simple here; every edge
belongs to the blob of the node it originates from. 

\paragraph{Primitives.} 
For primitive examples (e.g.~\word{touch} \cref{ex:prim_touch}) we mostly follow the same procedure. Unlike non-primitives, however, their resulting graph \emph{can} have open sources beyond the root node,
e.g.~\word{touch} would have sources at the nodes $b$ and $a$ (incoming \sortof{agent} or \sortof{theme} edge respectively).
These nodes can receive any source out of the three available (\srcr{S0},\srcr{S1},\srcr{S2})\footnote{with the restriction that different nodes should have different sources to prevent the nodes from being merged. Also we don't consider non-empty type requests for these nodes here.}, so the tree automaton build as part of \citet{groschwitz-etal-2021-learning}'s method would allow any combination of source names for the unfilled \sortof{arguments}.
Because there is only one input token, the alignment is trivial. 
In fact, primitives quite closely resemble the \sortof{supertags} of the AM parser.\\

Note that by encoding the logical form as a graph we get rid of the ordering of the conjuncts. The \sortof{correct} order (crucial for exact match evaluation) is restored during postprocessing.

The graph conversion for \cref{ex:boywanted} was already presented in \cref{fig:cogsgraph-boy}.
For the other three examples \crefrange{ex:other}{ex:prim_touch}, we present the graph conversions in \cref{fig:moreexamples}.

\end{document}